%% file: main.tex
\documentclass[letterpaper, 10 pt, conference]{ieeeconf}  

\IEEEoverridecommandlockouts                              

\pdfoutput=1

\usepackage{graphicx}
\usepackage{mathtools}
\usepackage{mathrsfs}
\usepackage{amssymb,amsfonts}

\usepackage{stmaryrd}
\usepackage[dvipsnames]{xcolor}
\usepackage[ruled,linesnumbered,vlined]{algorithm2e}
\usepackage[pdfusetitle]{hyperref}
\usepackage{booktabs}
\usepackage{siunitx}

\include{parts/macros}

\renewcommand{\leq}{\leqslant}
\renewcommand{\geq}{\geqslant}

\newcommand{\defeq}{ \overset{\text{def}}{=} }

\newcommand{\bmnu}{\bm{\nu}}

\newcommand{\mueq}{\mu_\mathrm{e}}
\newcommand{\muin}{\mu_\mathrm{i}}

\title{\LARGE \bf
    Constrained Differential Dynamic Programming:\\
    A primal-dual augmented Lagrangian approach
}

\author{%
    Wilson Jallet\textsuperscript{a,b,*},~%
    Antoine Bambade\textsuperscript{b,c},~%
    Nicolas Mansard\textsuperscript{a}~and~%
    Justin Carpentier\textsuperscript{b}
    \thanks{\textsuperscript{a}~LAAS-CNRS, 7 Avenue du Colonel Roche, F-31400 Toulouse, France}%
    \thanks{\textsuperscript{b}~Inria, Département d’informatique de l’ENS, \'Ecole normale supérieure, CNRS, PSL Research University, Paris, France}%
    \thanks{\textsuperscript{c}~ENPC, France,~\textsuperscript{*}{corresponding author}:
    \href{mailto:wjallet@laas.fr}{wjallet@laas.fr}}%
    \thanks{
        This work was supported in part by the HPC resources from
        GENCI-IDRIS (Grant AD011011342), the French government
        under management of Agence Nationale de la Recherche
        as part of the ”Investissements d’avenir” program, reference
        ANR-19-P3IA-0001 (PRAIRIE 3IA Institute) and ANR-19-
        P3IA-000 (ANITI 3IA Institute), Louis Vuitton ENS Chair
        on Artificial Intelligence, and the European project MEMMO
        (Grant 780684).
    }
}

\begin{document}


\maketitle
\thispagestyle{empty}
\pagestyle{empty}

\begin{abstract}
Trajectory optimization is an efficient approach for solving optimal control problems for complex robotic systems. 
It relies on two key components: first the transcription into a sparse nonlinear program, and second the corresponding solver to iteratively compute its solution. 
On one hand, differential dynamic programming (DDP) provides an efficient approach to transcribe the optimal control problem into a finite-dimensional problem while optimally exploiting the sparsity induced by time. 
On the other hand, augmented Lagrangian methods make it possible to formulate efficient algorithms with advanced constraint-satisfaction strategies.
In this paper, we propose to combine these two approaches into an efficient optimal control algorithm accepting both equality and inequality constraints.
Based on the augmented Lagrangian literature, we first derive a generic primal-dual augmented Lagrangian strategy for nonlinear problems with equality and inequality constraints.
We then apply it to the dynamic programming principle to solve the value-greedy optimization problems inherent to the backward pass of DDP, which we combine with a dedicated globalization strategy, resulting in a Newton-like algorithm for solving constrained trajectory optimization problems.
Contrary to previous attempts of formulating an augmented Lagrangian version of DDP, our approach exhibits adequate convergence properties without any switch in strategies.
We empirically demonstrate its interest with several case-studies from the robotics literature.

\end{abstract}

\input{parts/introduction}

\input{parts/pdal_method}
\input{parts/ddp}

\input{parts/experiments}

\input{parts/conclusion}

{
\footnotesize
\bibliographystyle{IEEEtran}
\bibliography{references}
}

\end{document}

%% file: parts/macros.tex
\usepackage{bm}
\usepackage{mathrsfs}

\newcommand{\RR}{\mathbb{R}}

\newcommand{\NN}{\mathbb{N}}

\newcommand{\calA}{\mathcal{A}}

\newcommand{\calK}{\mathcal{K}}
\newcommand{\calL}{\mathcal{L}}
\newcommand{\calM}{\mathcal{M}}

\newcommand{\calU}{\mathcal{U}}

\newcommand{\calX}{\mathcal{X}}

\newcommand{\suchthat}{\text{s.t.}\ }

\newcommand{\rmf}{\mathrm{f}}

\newcommand{\bfm}[1]{\mathbf{#1}}
\newcommand{\bfx}{\bfm{x}}
\newcommand{\bfu}{\bfm{u}}
\newcommand{\bmlam}{\bm{\lambda}}

\DeclareMathOperator{\proj}{proj}

\DeclareMathOperator*{\argmin}{arg\,min}

%% file: parts/introduction.tex
\section{Introduction}
\label{sec:introduction}

In this paper, we are interested in solving constrained continuous-time optimal control problems (OCP) of the form:
\begin{subequations}\label{eq:ContinuousOCP}
	\begin{alignat}{2}
		\min_{x, u}& \int_0^T \ell(t, x(t), u(t))\, dt + \ell_T(x(T)) \\ 
		\suchthat   & f(t, x(t), u(t), \dot{x}(t))= 0, \ t\in[0,T) \label{eq:ODE}  \\
		& x(0) = \bar{x}_0 \\
		& h(t, x(t), u(t)) \leq 0 \label{eq:COCP_path_constraints} \\
		& h_T(x(T)) \leq 0 \label{eq:COCP_terminal_constraints},
	\end{alignat}
\end{subequations}
where $\ell$ and $\ell_T$ are the running and terminal costs respectively, \eqref{eq:ODE} accounts for the system dynamics written as a differential-algebraic equation (DAE) (and includes the ODE case \mbox{$\dot{x} = f(t, x(t), u(t))$}).
We denote $\calX$ and $\calU$ the state and control spaces, $T>0$ the time horizon, $\bar{x}_0\in\calX$ the initial condition,
$h(\cdot)$ and $h_T(\cdot)$ the path and terminal constraints.

For numerical resolution, the continuous-time OCP~\eqref{eq:ContinuousOCP} must be transcribed into a finite-dimensional optimization problem (i.e., with a finite number of variables, which the continuous-time trajectories are not)~\cite{betts2005discretize}.
Several transcriptions are possible~\cite{hargraves1987direct,diehl2006fast,diehlEfficientNumericalMethods2009}.
Differential Dynamic Programming (DDP) is a particular OC algorithm which implies a direct transcription known as single shooting~\cite{murrayDifferentialDynamicProgramming1984}.
Popularized in robotics in the late 2000s~\cite{tassaSynthesisStabilizationComplex2012}, it has the advantage over other transcriptions of providing a simple formulation, optimally exploiting the  sparsity of the resulting nonlinear programs while providing feedback gains at no extra cost. 
The corresponding transcription, extended to any constraints, reads:
\begin{subequations}\label{eq:DiscreteOCP}
	\begin{alignat}{2}
		\min_{\bfx, \bfu}	& \sum_{k=0}^{N-1} \ell_k(x_k, u_k) + \ell_N(x_N)  \\
		\suchthat	& f_k(x_k, u_k, x_{k+1}) = 0, \ k\in\llbracket 0, N-1\rrbracket  \label{eq:dOCP_dyn} \\
		& x_0 = \bar{x}_0 \label{eq:dOCP_init} \\
		& h_k(x_k, u_k) \leq 0 \label{eq:dOCP_ineq} \\
		& h_N(x_N) \leq 0, \label{eq:dOCP_term}
	\end{alignat}
\end{subequations}
where $h_k, h_N, f_k$ are appropriate functions discretizing the dynamics and path constraints depending on the given numerical discretization scheme employed.
The $\ell_k$ are approximations of the cost integrals $\int_{t_k}^{t_{k+1}} \ell(t, x(t), u(t))\, dt$.
We use the shorthands \mbox{$\bfx \defeq (x_0, \ldots, x_N)$} and \mbox{$\bfu \defeq (u_0, \ldots, u_{N-1})$} for the discretized state and control trajectories.

\smallskip
While the nominal DDP algorithm is not able to handle path constraints, implicit integrators or multiple-shooting stabilization, several improvements have been proposed over the years to equip it with these properties. 
In this paper, we focus on the handling of equality and inequality constraints, and we first review previous work focusing on it.

A first subcase of interest only considers OCP with control bounds, which can be handled by a projected 
quasi-Newton approach~\cite{tassaControllimitedDifferentialDynamic2014}.
Several other projection-based formulations have then been proposed to extend DDP~\cite{giftthaler2017projection,xie2017differential}, none of which have been shown to be robust enough to be widely adopted in robotics.
To account fro inequality constraints, interior-point methods~\cite{pavlovInteriorPointDifferential2020,singhOptimizingTrajectoriesClosedLoop2021} have also been recently investigated; however, these do not allow for easy warm-starting~\cite{diehl2006fast} which is unsuitable for online optimization and application to model-predictive control (MPC)~\cite{wang2009fast,rawlings2017model}.

In the past few years, augmented Lagrangian approaches have emerged as a suitable solution for solving constrained trajectory optimization problems~\cite{toussaint2014novel}. 
As argued later in this paper, it offers many of the good properties that we need for trajectory optimization: super-linear convergence or even more quadratic convergence, stability, ability to warm-start, and so on.
Yet the first attempt to write dedicated OCP solvers based on augmented Lagrangians exhibited poor convergence properties. 
Thereby, further refinement using a projection in a two-stage approach had to be introduced in the solver ALTRO~\cite{howellALTROFastSolver2019}.
The penalty function used in ALTRO was then recognized to be irregular and discarded in~\cite{aoyamaConstrainedDifferentialDynamic2020}, which introduces a switch to an SQP formulation to converge to a higher precision.

A key idea that we exploit in this paper is to introduce the augmented Lagrangian formulation directly in the backward pass, to solve the value-greedy problems while directly considering the constraints, as initially proposed for multi-phase constrained problems~\cite{lantoineHybridDifferentialDynamic2013}.
This enables us to obtain better numerical accuracy for equality-constrained problems, by stabilizing the backward pass using a primal-dual system of equations to compute the control and multipliers together~\cite{kazdadiEqualityConstrainedDifferential2021}, and a monotonic update of the penalty parameter derived from the bound-constrained Lagragian (BCL)~\cite{connGloballyConvergentAugmented1991} strategy. Their method converges reliably to good numerical accuracy.
We have recently extended this formulation to also account for the dynamics and other equality constraints using a primal-dual augmented Lagrangian, allowing for the inclusion of infeasible initialization and implicit integrators~\cite{jalletImplicitDifferentialDynamic2022}.

\smallskip
In this paper, we introduce a complete augmented Lagrangian DDP algorithm for handling both equality and inequality constraints, and validate it on several real-size robotic scenarios. 
We first introduce in Sec.~\ref{sec:augmented_lagrangian} a primal-dual algorithm, rooted in the nonlinear programming literature~\cite{gillPrimaldualAugmentedLagrangian2012}, to handle generic nonlinear optimization problems (NLPs).
We then adapt it to the specific case of OCPs of the form~\eqref{eq:DiscreteOCP} in Sec.~\ref{sec:pdAL_ddp}. 
It results in an overall second-order quasi-Newton-like algorithm with good convergence properties for solving constrained trajectory optimization problems.
We finally benchmark our method in~Sec.\,\ref{sec:experiments} on various standard case studies from the robotics literature.
A companion video is available%
\footnote{\tiny\mbox{\url{https://peertube.laas.fr/videos/watch/dfeca51c-c2cf-468a-b46a-86f808e9a561}}}.

%% file: parts/pdal_method.tex
\section{The primal-dual augmented Lagrangian method for constrained optimization}
\label{sec:augmented_lagrangian}

This section introduces our augmented Lagrangian approach to solve constrained nonlinear optimization problems (NLP) of the form:
\begin{equation}\label{eq:ineqNLP}
\begin{alignedat}{2}
	\min_{x\in \mathbb{R}^n} &~ f(x)  \\
	\suchthat	& c(x) = 0,~ h(x) \leq 0,
\end{alignedat}
\end{equation}
where $c$ and $h$ stands for equality and inequality constraints respectively.
We then adapt this approach in~Sec.\,\ref{sec:pdAL_ddp} to the case of trajectory optimization.
While many augmented Lagrangian approaches have been introduced in the optimization literature~\cite{nocedalNumericalOptimization2006}, most of them rely on alternating between primal solving and dual updates.
In this work, we propose instead to compute combined primal-dual steps by taking inspiration from the work of Gill and Robinson in~\cite{gillPrimaldualAugmentedLagrangian2012}, which we extend by also considering inequality constraints and by connecting it to the proximal method of multipliers (PMM)~\cite{rockafellarAugmentedLagrangiansApplications1976} that we use to for numerical robustness. 
We discuss these contributions in more detail at the end of this section.

\subsection{Optimality conditions}

The Lagrangian $\mathcal{L}$ associated with~\eqref{eq:ineqNLP} is defined by: 
\begin{equation}\label{eq:lagrangian}
	\calL(x, \lambda, \nu) = f(x) + \lambda^\top c(x) + \nu^\top h(x), \
		\lambda \in \RR^{n_e}, \nu \in\RR_+^{n_i}
\end{equation}
A saddle point of $\mathcal{L}$ is a solution of \eqref{eq:ineqNLP}. 
This leads to the Karush-Kuhn-Tucker (KKT) necessary conditions~\cite{boyd2004convex} for ensuring a primal-dual point $(x,\lambda,\nu)$ to be optimal:
\begin{equation}\label{nlp:kkt}\tag{KKT}
\begin{aligned}
    \nabla_x \calL(x, \lambda, \nu) = 0,  \\
                            c(x)    = 0 \text{ and } h(x) \leq 0,  \\
                            \nu  \geq 0 \text{ and } h(x)^\top \nu = 0.
\end{aligned}
\end{equation}
In practice, we search for a triplet $(x,\lambda,\nu)$ satisfying these optimality conditions~\eqref{nlp:kkt} up to a certain level of predefined accuracy $\epsilon_{\text{abs}}>0$, leading us to the following natural absolute \textit{stopping criterion}:
\begin{equation}
\begin{aligned}\label{nlp:approx_nlp_sol}
    \left\{
    \begin{array}{l}
          \|\nabla_x \calL(x, \lambda, \nu)\|_{\infty} \leq  \epsilon_{\text{abs}},\\
         \|(c(x), [h(x)]_+)\|_{\infty} \leq \epsilon_{\text{abs}},
    \end{array}
    \right.
\end{aligned}
\end{equation}
where $[z]_+$ denotes the projection of $z\in\mathbb{R}^{n_z}$ on $\mathbb{R}_+^{n_z}$.

\subsection{Equality constrained nonlinear programming problems}

In this section, we provide a high-level overview on the primal-dual augmented Lagrangian~(PDAL) method. 
It is closely related to the probably even more famous Method of Multipliers (MM)~\cite[Chapter 2]{BertsekasBook}, which we review first in the context of purely equality-constrained NLPs.

\vspace{0.2cm}
\noindent\textbf{Primal-dual augmented Lagrangian.} 
The PDAL function $\calL^A_{\mu}$~\cite[Section 3]{gillPrimaldualAugmentedLagrangian2012} is defined by augmenting the standard Lagrangian $\mathcal{L}$~\eqref{eq:lagrangian} with two squared $\ell_2$ penalties:
\begin{equation}\label{eq:alm:eq}
\begin{split}
    \begin{alignedat}{2}
	\calM_{\mu}^A(x, \lambda;\lambda_e) &\defeq \mathcal{L}(x,\lambda_e,0) + \tfrac{1}{2\mueq}\| c(x) \|_2^2 \\
	 &+  \tfrac{1}{2\mueq}\|c(x)+\mu_e(\lambda_e - \lambda)\|_2^2,
	\end{alignedat}
\end{split}
\end{equation}
where $\mueq > 0$ is a scalar\footnote{Some authors associate a penalty parameter to each constraint. In this case, the penalty parameters $\mueq$ is a matrix $\Sigma_\lambda$~\cite{hermansQPALMNewtontypeProximal2019,hermansQPALMProximalAugmented2020}.}. 
The PDAL method then searches a sequence of iterates approximately minimizing \eqref{eq:alm:eq}~\cite{robinsonPrimaldualMethodsNonlinear2007}:
\begin{equation}\label{eq:generic:pdal}
    x_{l+1},\lambda_{l+1}\approx_{\omega_l}\min_{x,\lambda} \calM_{\mu}^A(x,\lambda;\lambda_l),
\end{equation}
where ${\approx}_{\omega_{l}}$ stands for requiring $(x_{l+1},\lambda_{l+1})$ to be an \mbox{$\omega_{l}$-approximate} solution to subproblem~\eqref{eq:generic:pdal}.
The approximation is controlled via the following condition:
\begin{equation}\label{eq:pdalm:termination}
    \|r_l(x_{l+1},\lambda_{l+1})\|_{\infty}\leq \omega_l,
\end{equation}
where $r_l$ accounts for the optimality conditions at iteration $l$: if $\|r_l(x,\lambda)\|_\infty \leq \epsilon_{\text{abs}}$ then $(x,\lambda)$ is a solution to~\eqref{eq:generic:pdal} at precision $\epsilon_{\text{abs}}$. The formula for $r_l$ will be specified in the sequel. The first subproblems are solved coarsely, since a precise solution is not required when the multiplier estimates $\lambda^0$ are far from the optimal dual solution: this avoids unnecessary computation.

To enforce the generated sequence $(x_l,\lambda_l)$ to converge to a local solution of NLP problem~\eqref{eq:ineqNLP}, we must address two important aspects: 
(i)~computing suitable iterates $(x_{l+1},\lambda_{l+1})$ satisfying~\eqref{eq:pdalm:termination} efficiently; (ii)~choosing appropriate rules for scheduling~$\omega_l$ ($\omega_l$ should decrease) and adequately increasing $\mueq$ (as shown by the theory, $\mueq$ should be increased over the iterations, but a too large value may drastically impact the overall numerical stability~\cite{nocedalNumericalOptimization2006}).

In the last two paragraphs of section~\ref{pdal:newton_semismooth:equality}, the problem (i) of finding suitable approximations for the subproblems is handled in the next paragraph. Then we review in the last paragraph the BCL globalization strategy (originating from~\cite{BCL}) for dealing with (ii). 

\vspace{0.2cm}
\noindent\textbf{Primal-dual Newton descent.}
\label{pdal:newton_semismooth:equality}
The stationarity condition of~\eqref{eq:generic:pdal} is	: 
\begin{equation}\label{eq:naive:kkt}
    \nabla\calM_{\mu}^A(x,\lambda;\lambda_l) = 0.
\end{equation}
Iterates $(x_{l+1},\lambda_{l+1})$ satisfying~\eqref{eq:naive:kkt} at precision $\omega_l$ can be derived using a quasi-Newton descent~\cite{gillPrimaldualAugmentedLagrangian2012}, which iteration at $t+1$ starting from $(\hat{x}_l^0,\hat{\lambda}_l^0)=(x_l,\lambda_l)$ reads:
\begin{equation}\label{eq:semi_smooth_newton_system}
\begin{aligned}
    &\begin{bmatrix}
    H_l +\frac{1}{\mu_e}J_c^\top J_c & - J_c^\top\\
    -J_c & \mu_e I
    \end{bmatrix}
    \begin{bmatrix}
    \delta x\\\delta\lambda
    \end{bmatrix} = -\begin{bmatrix}
    \nabla_x \mathcal{L}^A_{\mu}(\hat{x}_l^t,\hat{\lambda}_l^t;\lambda_l) \\
    \nabla_{\lambda} \mathcal{L}^A_{\mu}(\hat{x}_l^t,\hat{\lambda}_l^t;\lambda_l)
    \end{bmatrix},
\end{aligned}
\end{equation}
with:
\begin{equation}
    \hat{x}_l^{t+1} = \hat{x}_l^{t}+\delta x,\quad
    \hat{\lambda}_l^{t+1} = \hat{\lambda}_l^{t}+\delta \lambda,
\end{equation}
and where $H_l$ is the Lagrangian Hessian $\nabla_x^2\calL(\hat{x}_l^t, 2\pi_l(\hat{x}_l^t)-\hat{\lambda}_l^t,0)$ or an approximation thereof, with $\pi_l(x)\defeq \lambda_l + \tfrac{1}{\mueq}c(x)$ (following the notation in~\cite{gillPrimaldualAugmentedLagrangian2012}), and $J_c$ the constraint Jacobian matrix at $\hat{x}_l^t$.

There are two conflicting goals to balance in this iterative process. 
First, the smaller the value of $\mueq$ the faster the convergence, as $\tfrac{1}{\mu_e}$ penalizes the constraints. Second, as $\mueq$ gets smaller, the conditioning of $(H_l+\tfrac{1}{\mueq}J_c^TJ_c)$ gets worse, thereby harming the numerical robustness of the approach, in particular when $J_c$ has a large condition number.

Fortunately, the linear system~\eqref{eq:semi_smooth_newton_system} can be rewritten in the following equivalent form:
\begin{equation}\label{equality:primaldualNewton}
	\begin{bmatrix}
		H_l	& J_c^\top \\
		J_c	& -\mueq I \\
	\end{bmatrix}
	\begin{bmatrix}
		\delta x \\ \delta\lambda \\
	\end{bmatrix}
	=
	-\begin{bmatrix}
		\nabla_x \calL(\hat{x}_l^t, \hat{\lambda}_l^t, 0)  \\
		c(\hat{x}_l^t) + \mueq(\lambda_l - \hat{\lambda}_l^t), \\
	\end{bmatrix},
\end{equation}
which shows that the PDAL method is closely related to the method of multipliers (MM)~\cite[Chapter 2]{BertsekasBook}. Indeed,~\eqref{equality:primaldualNewton} implies that the sequence of iterates $(x_l,\lambda_l)$ approximate those of a proximal-point method applied to the dual of~\eqref{eq:lagrangian}:
\begin{equation}\label{eq:mm}
	x_{l+1},\lambda_{l+1}\approx_{\omega_l}\min_x\max_{\lambda}
	\calL(x, \lambda, 0) - \tfrac{\mueq}{2} \|\lambda - \lambda_l\|_2^2.
\end{equation}
Hence, $\mueq$ is the inverse of the step-size of the equivalent MM, and it directly calibrates the convergence speed of the approach (see~\cite[Section 4]{rockafellarAugmentedLagrangiansApplications1976} for details). 
Moreover, this linear system involves a matrix that is always nonsingular thanks to the regularization terms $-\tfrac{\mueq}{2}\|\lambda-\lambda_l\|^2_2$. 
In other words, the problem~\eqref{equality:primaldualNewton} is always well-defined in the iterative process. Such linear systems are also better conditioned than~\eqref{eq:semi_smooth_newton_system}~\cite[Section 17.1]{nocedalNumericalOptimization2006}. 

Finally, \eqref{equality:primaldualNewton} implies that the sequence $(\hat{x}^t_l, \hat{\lambda}^t_l)_{t\geq 0}$ converges to a pair $(x^{*},\lambda^{*})$ satisfying the following optimality conditions:
\begin{equation}\label{pdal:optimality}
    \begin{bmatrix}
		\nabla_x \calL(x^{*}, \lambda^{*}, 0)  \\
		c(x^{*}) + \mueq(\lambda_l - \lambda^{*})  \\
	\end{bmatrix}=0.
\end{equation}
Hence, we choose the optimality criterion function $r_l$ to be:
\begin{equation}\label{in:r_l}
\begin{aligned}
    r_l(x,\lambda)\defeq\begin{bmatrix}
		\nabla_x \calL(x, \lambda, 0)  \\
		c(x) + \mueq(\lambda_l - \lambda)  \\
	\end{bmatrix}.
\end{aligned}
\end{equation}

\vspace{0.2cm}
\noindent\textbf{The globalization strategy.}
\label{subsec:bcl}
For fixing the hyper-parameters (tolerance on subproblems $\omega_l$, step-sizes $\mueq$), we rely on BCL (see~\cite{connGloballyConvergentAugmented1991} and~\cite[Algorithm 17.4]{nocedalNumericalOptimization2006}) which has been proved to perform well in advanced optimization packages such as LANCELOT~\cite{LANCELOT} and also in robotics for solving constrained optimal control problems~\cite{aoyamaConstrainedDifferentialDynamic2020,plancherConstrainedUnscentedDynamic2017}. 

The main idea underlying BCL consists in updating the dual variables $\lambda_l$ from~\eqref{eq:mm} only when the corresponding primal feasibility (denoted by $\eta_l$ hereafter) is small enough. More precisely, we use a second sequence of tolerances denoted by $\epsilon_l$ (which we also tune within the BCL strategy) and update the dual variables only when $\eta_{l+1}\leq\epsilon_l$, where $\eta_{l+1}$ denotes the primal infeasibility as follows:
\begin{equation}
\begin{aligned}
    \eta_{l+1}\defeq&\|c(x_{l+1})\|_{\infty}.
\end{aligned}
\end{equation}
It remains to explain how the BCL strategy chooses appropriate values for the hyper-parameters $\omega_l$, $\epsilon_l$ and $\mu_e$. As for the update of the dual variables, it proceeds in two stages:
\begin{itemize}
    \item {\bf{}If $\eta_{l+1}<\epsilon_{l}$:} the primal feasibility is good enough, we thus keep the constraint penalization parameters as is.  
    \item {\bf{}Otherwise:} the primal infeasibility is too large, we thus increase quadratic penalization terms on the constraints for the subsequent subproblem~\eqref{eq:generic:pdal}.
\end{itemize}
Concerning the accuracy parameters $\omega_l$ and $\epsilon_l$, the update rules are more technical and the motivation underlying those choices is to ensure global convergence: an exponential-decay type update when primal feasibility is good enough, and see~\cite[Lemma 4.1]{BCL} for when the infeasibility is too large. The detailed strategy is summarized in Algorithm~\ref{al:PDAL_BCL} for the general case (including inequalities).

\subsection{Extension to inequality constrained nonlinear programs}

As we will see, our approach developed for tackling equality constraints easily extends to the general case. Indeed, as we will see the PDAL function only changes in a subtle way for taking into account inequality constraints. As a result, it also impacts how the minimization procedure must be realized.

\vspace{0.2cm}
\noindent\textbf{Generalized primal-dual merit function.}
\label{subsec:generalized_pd_merit_func}
In the general setup, the PDAL function can be framed in its equality constrained form introducing a slack variable $z\leq0$ satisfying the new equality constraint:
\begin{equation}
    h(x)-z = 0.
\end{equation}
Hence, the generalized PDAL function reads:
\begin{equation}\label{generalized:alm}
\begin{split}
    \begin{alignedat}{3}
	\calL_{\mu}^A(x, \lambda, & \nu, z; \lambda_l, \nu_l) \defeq\mathcal{L}(x,\lambda_l,\nu_l) + \tfrac{1}{2\mueq}\| c(x) \|_2^2 \\
	 &+\tfrac{1}{2\mueq}\|c(x)+\mu_e(\lambda_l - \lambda)\|_2^2 + \tfrac{1}{2\muin}\| h(x)-z \|_2^2 \\
	 &+\tfrac{1}{2\muin}\|h(x)-z+\mu_i\nu_l -\mu_i \nu\|_2^2 + g(z).
	\end{alignedat}
\end{split}
\end{equation}
 $g$ is the (component-wise) indicator function related to $z\leq0$:
\begin{equation*}
g(z)\defeq\left\{\begin{array}{ll}
0   \quad  & \text{if } z_i\leq0, i\in[1,n_i],  \\
+\infty     & \text{otherwise.}
\end{array}\right.
\end{equation*}
The minimization of~\eqref{generalized:alm} w.r.t. $x$, $\lambda$, $\nu$ or $z$ variables commutes. 
Considering the problem structure and following ideas from~\cite{demarchiPrimaldualNewtonProximal2022}, it can be shown that $z$ and $\nu$ can be directly deduced as functions of $x$:
\begin{equation}\label{eq:nu_first_order:update}
\begin{aligned}
    \hat{z}(x,\nu_l),\hat{\nu}(x,\nu_l) &\defeq \argmin_{z,\nu} \calL_{\mu}^A(x,\lambda,\nu,z;\lambda_l,\nu_l),\\
    \hat{z}(x,\nu_l) &= [h(x)+ \muin\nu_l]_{-},\\
    \hat{\nu}(x,\nu_l) &= \left[\tfrac{1}{\muin}h(x)+ \nu_l \right]_+.
\end{aligned}
\end{equation}
The minimization problem can thus be reduced to:
\begin{equation}\label{generalized:merit_function}
\begin{split}
    \begin{alignedat}{3}
	&\min_{x,\lambda,\nu,z} \calL_{\mu}^A(x, \lambda, \nu, z; \lambda_l, \nu_l) \\
	&=\min_{x,\lambda} \calL_{\mu}^A(x, \lambda, \hat{\nu}(x,\nu_l), \hat{z}(x,\nu_l); \lambda_l, \nu_l).
	\end{alignedat}
\end{split}
\end{equation}
Yet, we choose to maintain the $\nu$ variable relaxed in the minimization procedure~\eqref{generalized:merit_function} as it enables us to preserve similar well conditioned linear systems and stopping criterion derived in~\eqref{equality:primaldualNewton} and~\eqref{pdal:optimality}. 
Consequently, the generalized merit function corresponds to:
\begin{equation}\label{final:pdALM_fun}
	\begin{split}
		&\calM_\mu(x, \lambda, \nu; \lambda_l, \nu_l) \defeq 
		\calL_{\mu}^A(x, \lambda, \nu, \hat{z}(x,\nu_l); \lambda_l, \nu_l) \\
		&= f(x) + \tfrac{1}{2\mueq}\|c(x) + \mueq\lambda_l\|_2^2+\tfrac{1}{2\mueq}\big\| c(x) + \mueq(\lambda_l - \lambda) \big\|_2^2\\
		&+ \tfrac{1}{2\muin} \big\| \left[  h(x) + \muin\nu_l \right]_+ \big\|^2_2
		+ \tfrac{1}{2\muin} \big\| \left[ h(x) + \muin\nu_l \right]_+ -  \muin\nu \big\|_2^2.
	\end{split}
\end{equation}
For ensuring better regularization w.r.t. the primal variable $x$, following the PMM~\cite{rockafellarAugmentedLagrangiansApplications1976}, we finally consider the following generalized primal-dual merit function:
\begin{equation}\label{final:prox:pdALM_fun}
	\calM_{\mu,\rho}(x, \lambda, \nu; x_l, \lambda_l, \nu_l)\defeq \calM_{\mu}(x, \lambda, \nu; \lambda_l, \nu_l)+\frac{\rho}{2}\|x- x_l\|_2^2,
\end{equation}
with $\rho>0$ a proximal parameter. 

\vspace{0.2cm}
\noindent\textbf{Semi-smooth Newton step with line-search procedure.}
\label{subsec:semismoothNewton}
Contrary to the equality-constrained case,~\eqref{final:prox:pdALM_fun} now corresponds to a semi-smooth function (due to the presence of the positive orthant projection operators $[\cdot]_+$).
It should thus be minimized using a semi-smooth quasi-Newton iterative procedure~\cite[Chapter 1]{hintermullerSemismoothNewtonMethods},~\cite[Chapter 6]{nocedalNumericalOptimization2006}. 
To ensure convergence, the procedure must have a line-search scheme\footnote{Different line-search schemes can be used to minimize the merit function~\eqref{final:prox:pdALM_fun} through a semi-smooth quasi-Newton iterative procedure.}  over the semi-smooth convex primal-dual merit function~\eqref{final:prox:pdALM_fun} to find an adequate step size along the primal-dual Newton direction, following an approach similar to~\eqref{equality:primaldualNewton}, while also considering the change of active inequality-constraints defined by $\calA_l(x)$:
\begin{equation}
\begin{aligned}
	\calA_l(x) \defeq \{ j  \mid (\nu_l + \muin h_j(x)) \geq 0\},
\end{aligned}
\end{equation}
where $\calA_l(x)$ is the shifted active-set of the $l$-th subproblem at point $x$. 
This definition of shifted active set differs from existing augmented Lagrangian-based optimal control methods in robotics which defines the active-set by the condition $h_j(x) \geq 0$, as done in~\cite{howellALTROFastSolver2019,li2020hybrid}.
%

\vspace*{-2mm}
\begin{algorithm}
\caption{PDAL Method for constrained optimization}\label{al:PDAL_BCL}
\SetAlgoLined
 \textbf{Inputs}:\\ 
 \begin{itemize}
     \item initial states: $x_0$, $\lambda_0$, $\nu_0$,
     \item initial parameters: $\epsilon_0,\, \omega_0,\, \rho,\, \mu_e,\, \mu_i>0$,
    \item hyper-parameters: $\mu_f<1,\alpha_{\text{bcl}}\in(0,1),\beta_{\text{bcl}}\in(0,1)$, $\underline{\muin},\underline{\mueq}>0$.
 \end{itemize}
 \While{Stopping criterion~\eqref{nlp:approx_nlp_sol} not satisfied}{
    Compute $(\widetilde{x}_{l+1},\widetilde{\lambda}_{l+1},\widetilde{\nu}_{l+1})$ satisfying~\eqref{generalized:inner_loop:exit} using~\ref{subsec:semismoothNewton}  \;
    $x_{l+1}= \widetilde{x}_{l+1}$\;
   \eIf{$\eta_{l+1}<\epsilon_l$}{
    $\epsilon_{l+1} = \epsilon_l\muin^{\beta_{\text{bcl}}}$; $\omega_{l+1}=
    \omega_{l} \muin$;
    
    $\lambda_{l+1}=2\hat{\lambda}(x_{l+1}, \lambda_l) - \widetilde{\lambda}_{l+1}$;
    
    $\nu_{l+1}= [2\hat{\nu}(x_{l+1}, \nu_l) - \widetilde{\nu}_{l+1}]_+$;
   }{
    $\muin \xleftarrow{} \max(\underline{\muin},\mu_{f}\muin), \mu_e\xleftarrow{}\max(\underline{\mueq},\mu_{f}\mueq)$;
    
    $\epsilon_{l+1} = \epsilon_0\muin^{\alpha_{\text{bcl}}}$; $\omega_{l+1} = \omega_0\muin$;
    
     $\lambda_{l+1}=\lambda_l$; $\nu_{l+1}= \nu_{l}$;
    }
    $l\leftarrow l+1$;
 }
 \textbf{Output}: A $(x_{l},\lambda_{l},\nu_{l})$ satisfying the $\epsilon_{\text{abs}}$-approximation criterion~\eqref{nlp:approx_nlp_sol} for problem~\eqref{eq:ineqNLP}.
\end{algorithm}
\vspace*{-5mm}

\subsection{Final algorithm}
\label{final:alg:nlp}

Once a local $\omega_l$ primal-dual solution $(x^*, \lambda^*, \nu^*)$ minimizing~\eqref{final:prox:pdALM_fun} is found, for better numerical precision, we follow the Lagrange multiplier update rule introduced in~\cite[Section 4]{robinsonPrimaldualMethodsNonlinear2007}:
\begin{equation}\label{eq:pdALM_mulUpdate}
\begin{aligned}
	\lambda_{l+1} &= 2\hat{\lambda}(x^*, \lambda_l) - \lambda^*,  \\
	\nu_{l+1} &= [2\hat{\nu}(x^*, \nu_l) - \nu^*]_+,
\end{aligned}
\end{equation}
where $\hat{\nu}(x,\nu_l)$ is defined from the derivation of our generalized merit function in~\eqref{eq:nu_first_order:update}, and $\hat{\lambda}(x, \lambda_l)$ comes from the classic multiplier update rule (similar to $\hat{\nu}(x,\nu_l)$ without projections)~\cite{robinsonPrimaldualMethodsNonlinear2007}. Hence, the measure of the convergence towards the optimality conditions captured by $r_l$~\eqref{in:r_l}
can be more generally defined as follows:
\begin{equation}\label{generalized:r_l}
    r_l(x,\lambda,\nu) \defeq    \begin{bmatrix}
        \nabla_x \calL(x, \lambda, \nu) + \rho (x - x_l) \\
        \mueq (\hat{\lambda}(x, \lambda_l) - \lambda)\\
        \muin (\hat{\nu}(x, \nu_l) - \nu)
    \end{bmatrix}.
\end{equation}
The generalized inner-loop exit condition thus reads:
\begin{equation}\label{generalized:inner_loop:exit}
    \|r_l(x,\lambda,\nu)\|_{\infty}\leq\omega_l.
\end{equation}
The primal feasibility also generalizes as:
\begin{equation}\label{eq:primalFeasibility}
\begin{aligned}
    \eta_{l+1}\defeq&\|(c(x_{l+1}),[h(x_{l+1})]_+)\|_{\infty}.
\end{aligned}
\end{equation}
Algorithm~\ref{al:PDAL_BCL} summarizes our approach for solving NLPs. 

\subsection{Key novelties of Algorithm~\ref{al:PDAL_BCL}}
Algorithm~\ref{al:PDAL_BCL} differs from~\cite{gillPrimaldualAugmentedLagrangian2012}
for two main aspects. 
First, we show in the equality-constrained case that the linear systems involved in the (quasi-)Newton steps are equivalent to a better-conditioned linear system originating from the proximal method of multipliers~\cite{rockafellarAugmentedLagrangiansApplications1976}. 
For this reason, we use this equivalent saddle-point system formulation and its associated stopping criterion to enforce the overall numerical stability of the approach. 
Second, we extend the PDAL function from~\cite{gillPrimaldualAugmentedLagrangian2012} to account for inequality constraints by introducing a new merit function that does not require any slack variables. 

The resulting algorithm is a generic NLP solver which is a contribution in itself, with direct application in optimization for robotics e.g.~\cite{brossette2018multicontact}. 
As our aim is to design a constrained OCP solver, we have left evaluation of this generic solver's performance for future work, and we directly jump to its adaptation to dynamic programming. 

%% file: parts/ddp.tex
\section{Primal-Dual Augmented Lagrangian for Constrained Differential Dynamic Programming}
\label{sec:pdAL_ddp}

In this section, we extend the differential dynamic programming framework accounting for equality constraints \cite{kazdadiEqualityConstrainedDifferential2021} and implicit dynamics~\cite{jalletImplicitDifferentialDynamic2022} to the case of inequality constraints using the PDAL introduced 
in Sec.~\ref{sec:augmented_lagrangian}.

\subsection{Relaxation of the Bellman equation.}

In \cite{jalletImplicitDifferentialDynamic2022}, we show that applying the MM to \eqref{eq:DiscreteOCP} leads to a relaxation of the Bellman equation, in the equality-constrained case. We now extend this idea to the inequality-constrained case.
Indeed, the discrete-time problem \eqref{eq:DiscreteOCP} also satisfies a dynamic programming equation in the inequality-constrained case. 
The value function for the subproblem at time $k$ satisfies the Bellman relation:
\begin{equation}\label{eq:BellmanEqn}
	\begin{alignedat}{2}
		V_k(x) &= \min_{u, x'} \ell_k(x, u) + V_{k+1}(x')  \\
		\suchthat & f_k(x, u, x') = 0 \text{ and }
		 h_k(x, u) \leq 0.
	\end{alignedat}
\end{equation}
The optimality conditions for this Bellman equation involve a Lagrangian function of the form:
\begin{equation}
    \begin{split}
        \calL_k(x,u,x',\lambda,\nu) &= \ell_k(x,u) + V_{k+1}(x') \\
        &+ \lambda^\top f_k(x,u,x') +  \nu^\top h_k(x,u).
    \end{split}
\end{equation}
The last term in this Lagrangian, relating to the inequality constraint, will appear in the KKT conditions of the Bellman equation and influence the sensitivities with respect to $x$.
Following the PDAL method from Sec.~\ref{sec:augmented_lagrangian}, we can define an augmented \textit{primal-dual $Q$-function} modelled after \eqref{final:pdALM_fun} which reads,
considering multiplier estimates $(\lambda_l, \nu_l)$, $l\in\NN$:
\begin{equation}
\hspace*{-0.5em}
\begin{split}
    &Q_{\mu,k}^l (x, u, x', \lambda, \nu) = \ell_k(x, u) + V_{k+1}(x')  \\
	&+ \tfrac{\mueq}{2}(\| \tfrac1\mueq f_k(x,u,x') + \lambda_l\|^2 + \| \tfrac{1}{\mueq}f_k(x, u, x') + \lambda_l-\lambda \|^2)  \\
	&+ \tfrac{\muin}{2}(\| [\nu_l+ \tfrac1\muin h_k(x, u)]_+\|^2 + \| [\nu_l + \tfrac1\muin h_k(x, u)]_+ - \nu\|^2).
\end{split}
\end{equation}
Then, the minimization in the Bellman equation is relaxed to the following augmented Lagrangian iteration:
\begin{equation}\label{eq:pdAL_Bellman}
	V_k^l(x) = \min_{u,x'\lambda,\nu} Q_{\mu,k}^l(x, u, x', \lambda, \nu),
\end{equation}
with the boundary condition $V^l(x) = \ell_\rmf(x)$.
This dynamic programming equation can be seen as a relaxation of the classical Bellman equation \eqref{eq:BellmanEqn} using the primal-dual merit function -- including the dynamics $f_k(x,u,x')$ in this relaxation leads to a multiple-shooting formulation.
Indeed, assuming the multipliers estimates $(\lambda_l, \nu_l)$ are optimal multipliers associated with \eqref{eq:BellmanEqn}, then any minimizer $(\Bar{u}, \Bar{x}', \Bar{\lambda}, \Bar{\nu})$ of \eqref{eq:pdAL_Bellman} also satisfies the optimality conditions for \eqref{eq:BellmanEqn}.

\begin{figure*}[ht!]
    \vspace{2mm}
    \centering
    \def\snapwidth{1.95cm}
    \def\snapleft{700pt}\def\snapright{560pt}
    \def\snapbot{200pt}
    \def\snaptop{220pt}%
    \def\snapleftB{560pt}\def\snaprightB{500pt}%
    \def\snapbotB{160pt}
    \def\snaptopB{60pt}
    \setlength{\tabcolsep}{0.1mm}
    \begin{tabular}{@{}llll|rrrrr}
        \includegraphics[trim=\snapleft{} \snapbot{} \snapright{} \snaptop{},clip=true,width=\snapwidth]{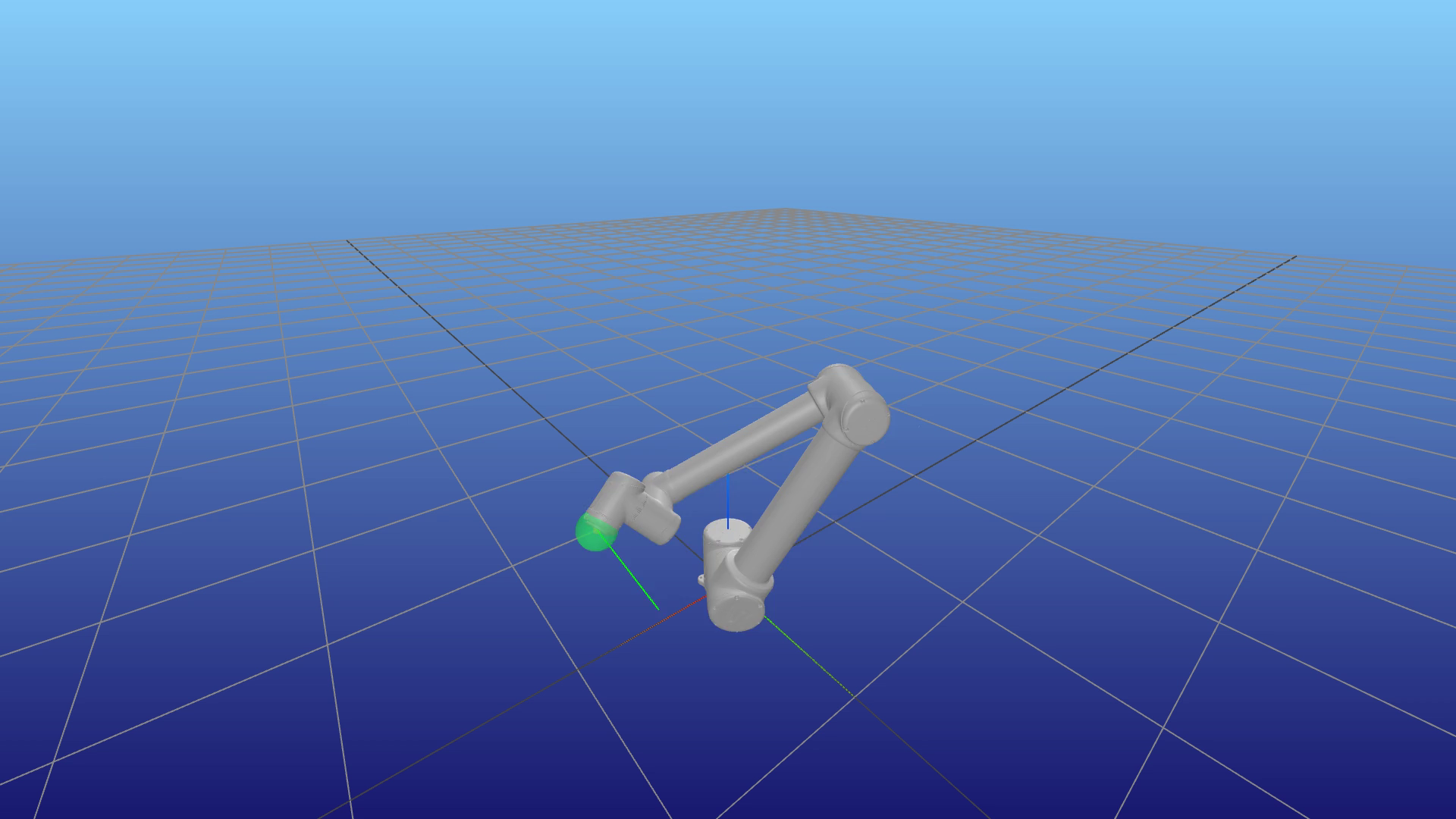} &
        \includegraphics[trim=\snapleft{} \snapbot{} \snapright{} \snaptop{},clip=true,width=\snapwidth]{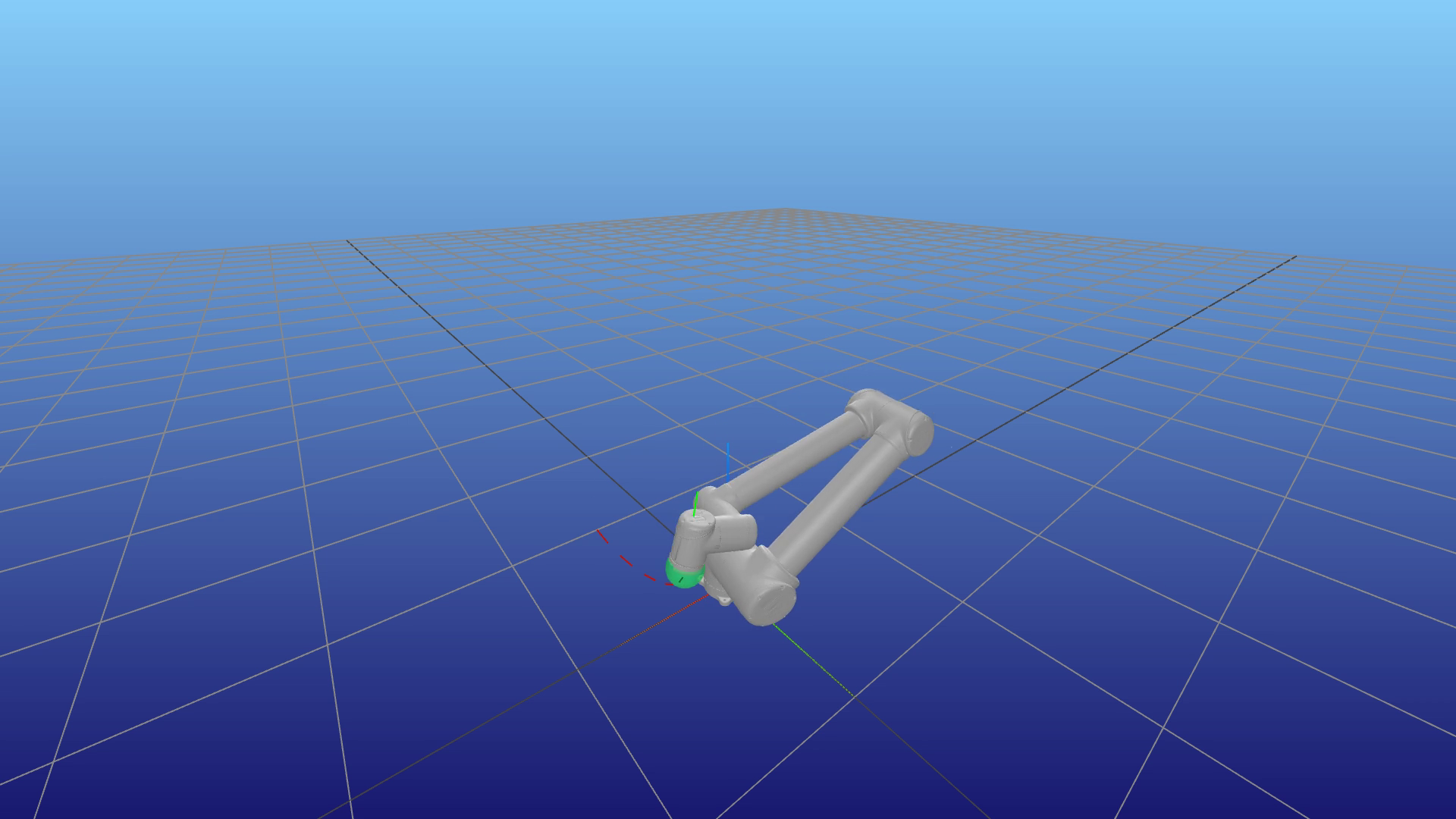} &
        \includegraphics[trim=\snapleft{} \snapbot{} \snapright{} \snaptop{},clip=true,width=\snapwidth]{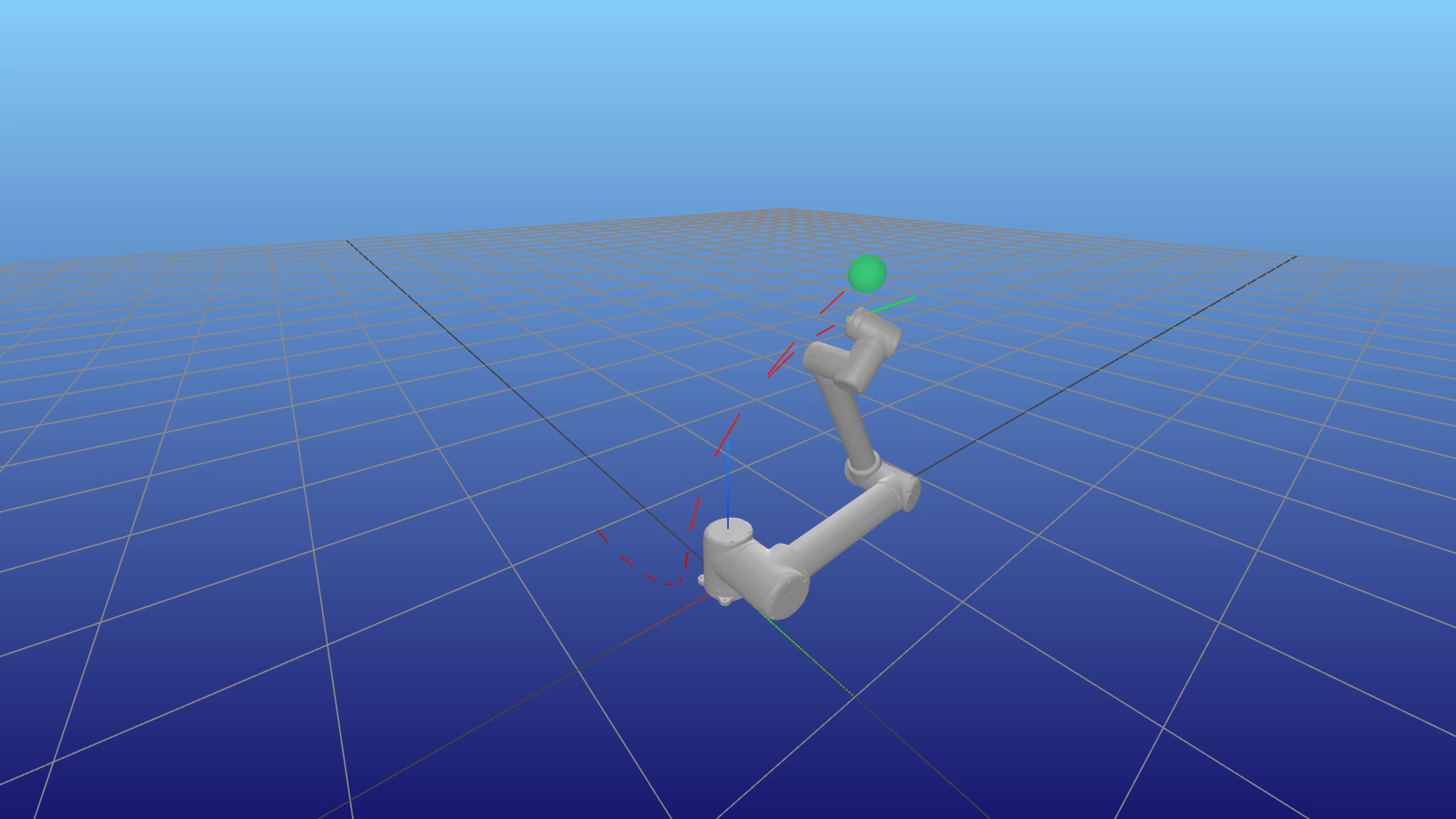} &
        \includegraphics[trim=\snapleft{} \snapbot{} \snapright{} \snaptop{},clip=true,width=\snapwidth]{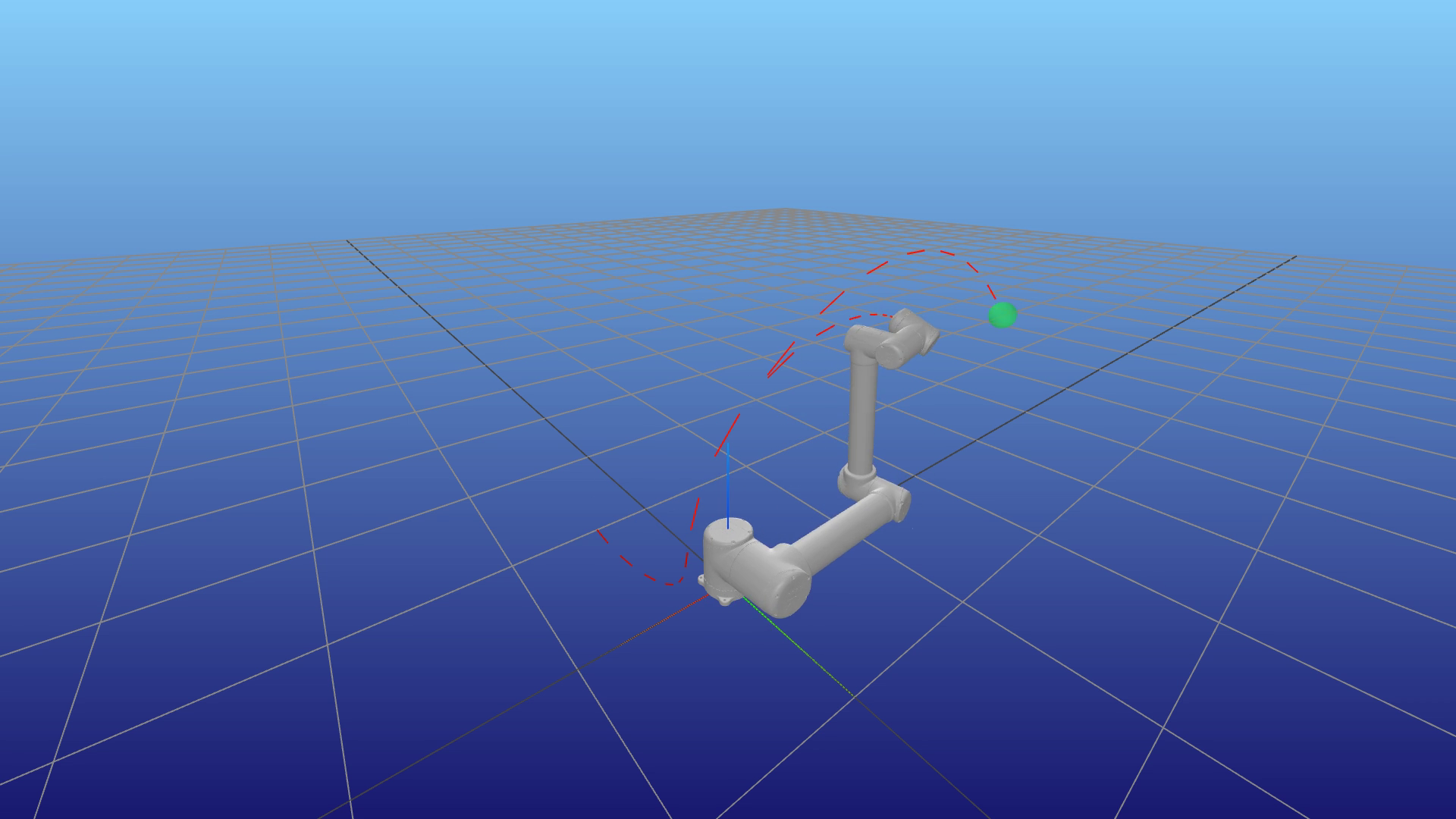} &
        \includegraphics[trim=\snapleftB{} \snapbotB{} \snaprightB{} \snaptopB{},clip=true,width=\snapwidth]{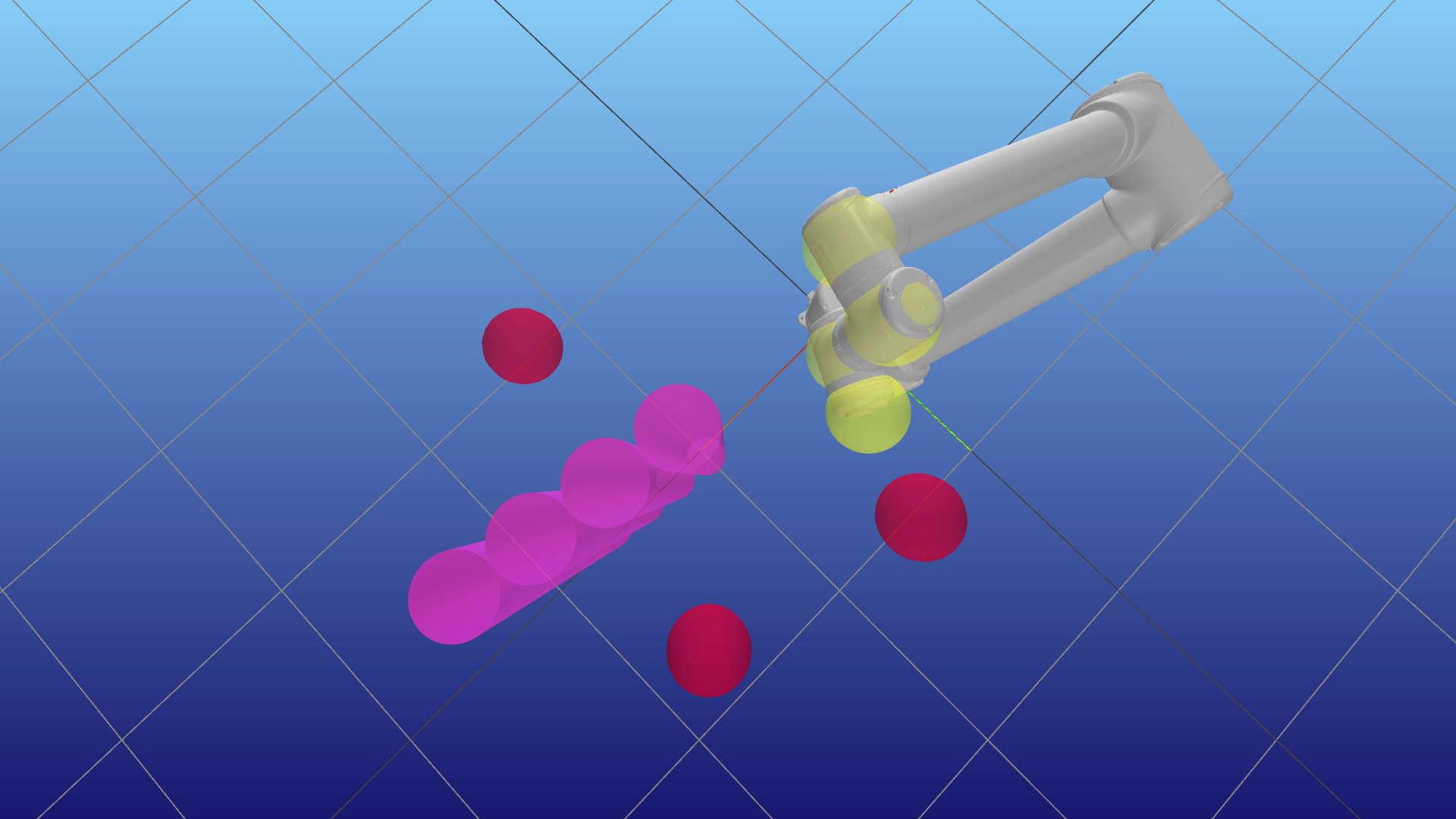} &
        \includegraphics[trim=\snapleftB{} \snapbotB{} \snaprightB{} \snaptopB{},clip=true,width=\snapwidth]{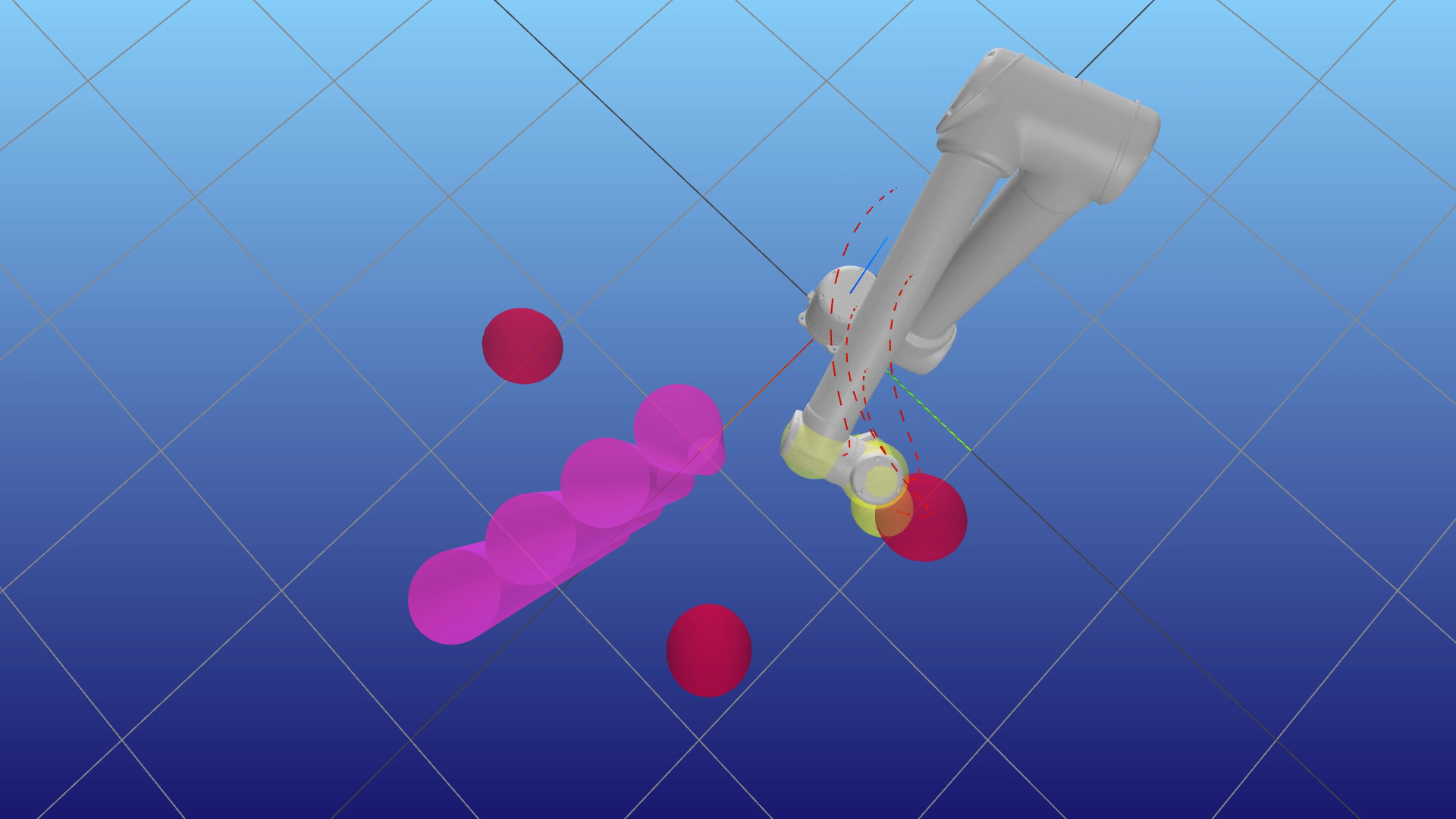} &
        \includegraphics[trim=\snapleftB{} \snapbotB{} \snaprightB{} \snaptopB{},clip=true,width=\snapwidth]{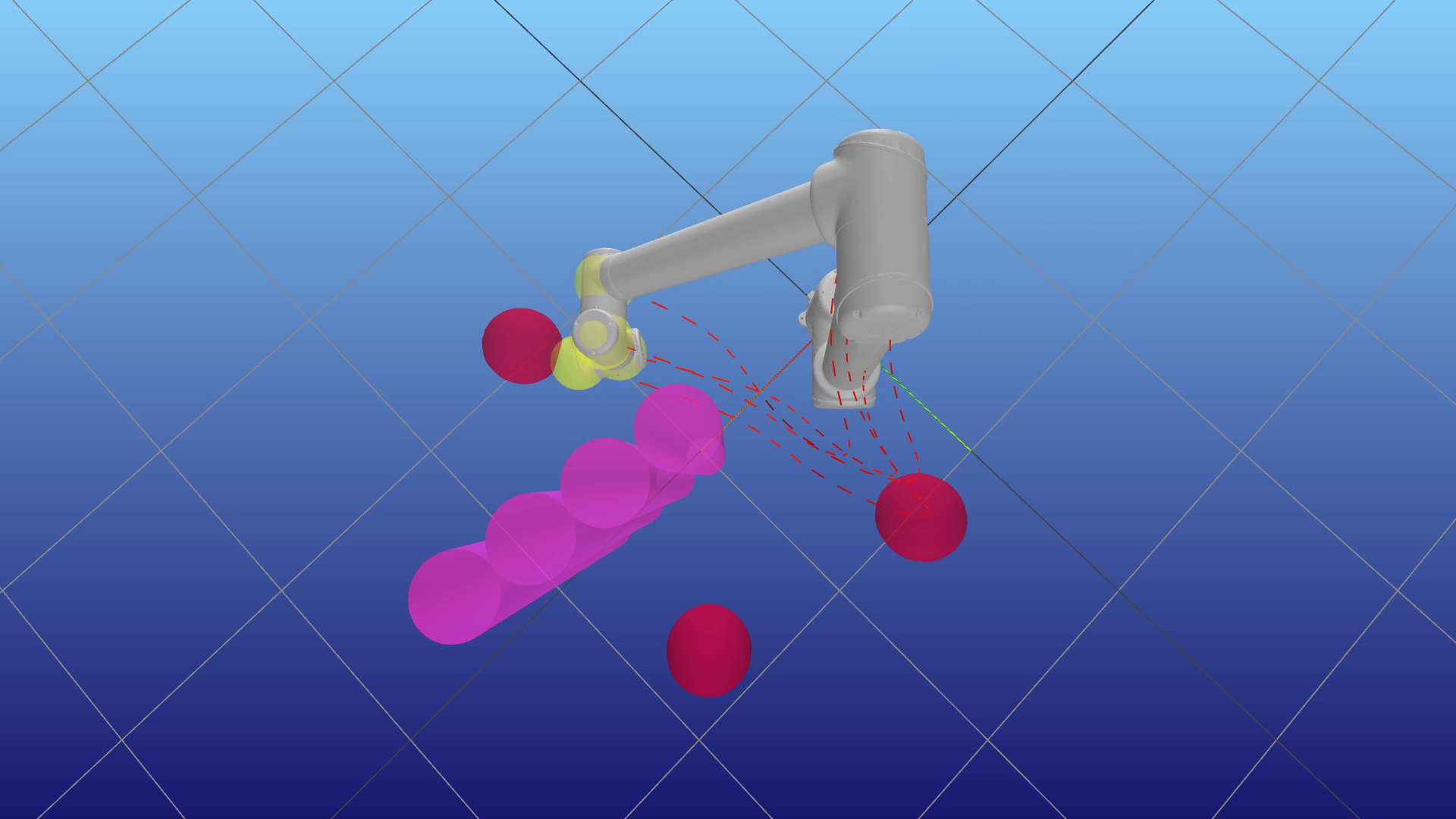} &
        \includegraphics[trim=\snapleftB{} \snapbotB{} \snaprightB{} \snaptopB{},clip=true,width=\snapwidth]{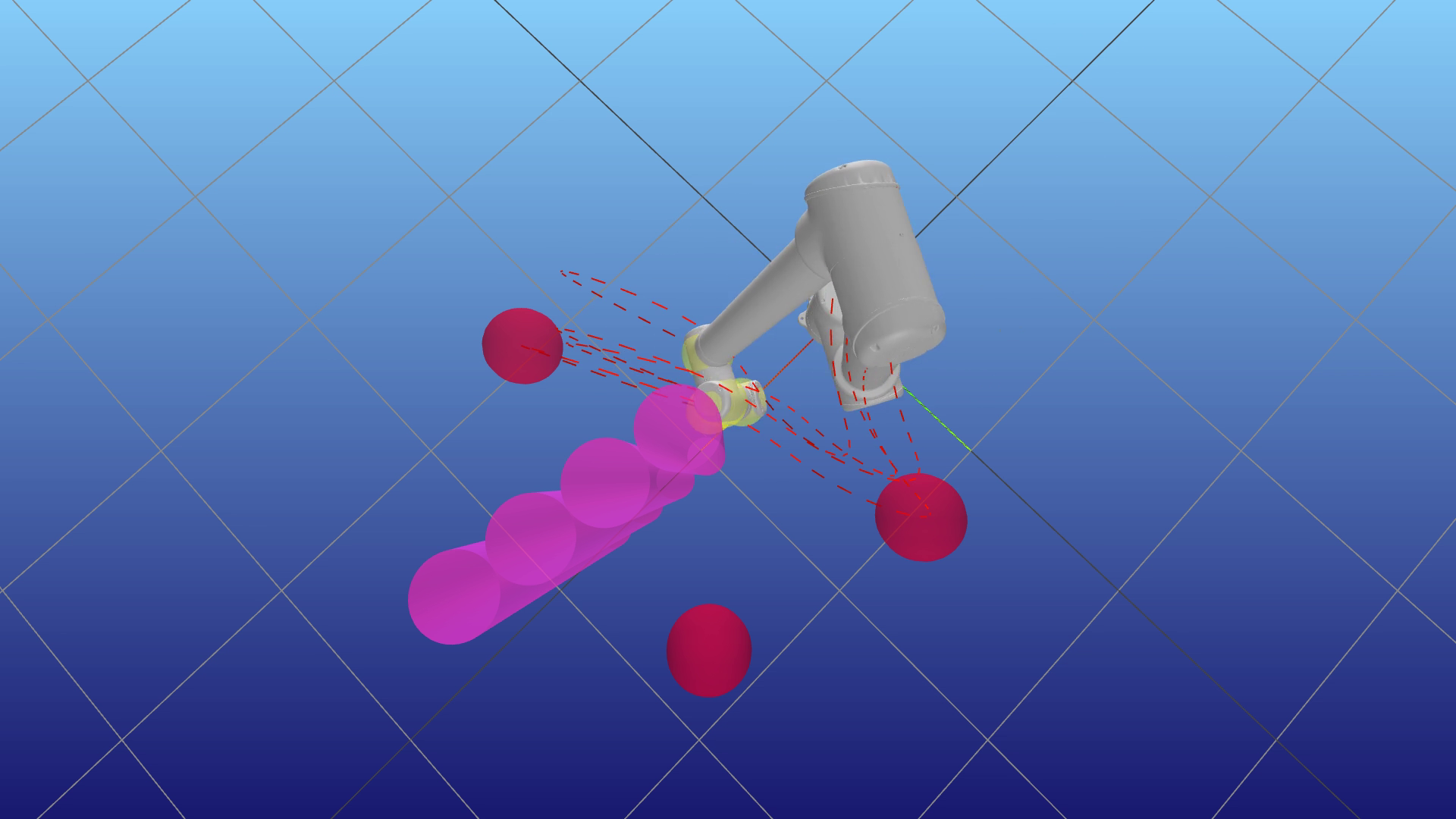} &
        \includegraphics[trim=\snapleftB{} \snapbotB{} \snaprightB{} \snaptopB{},clip=true,width=\snapwidth]{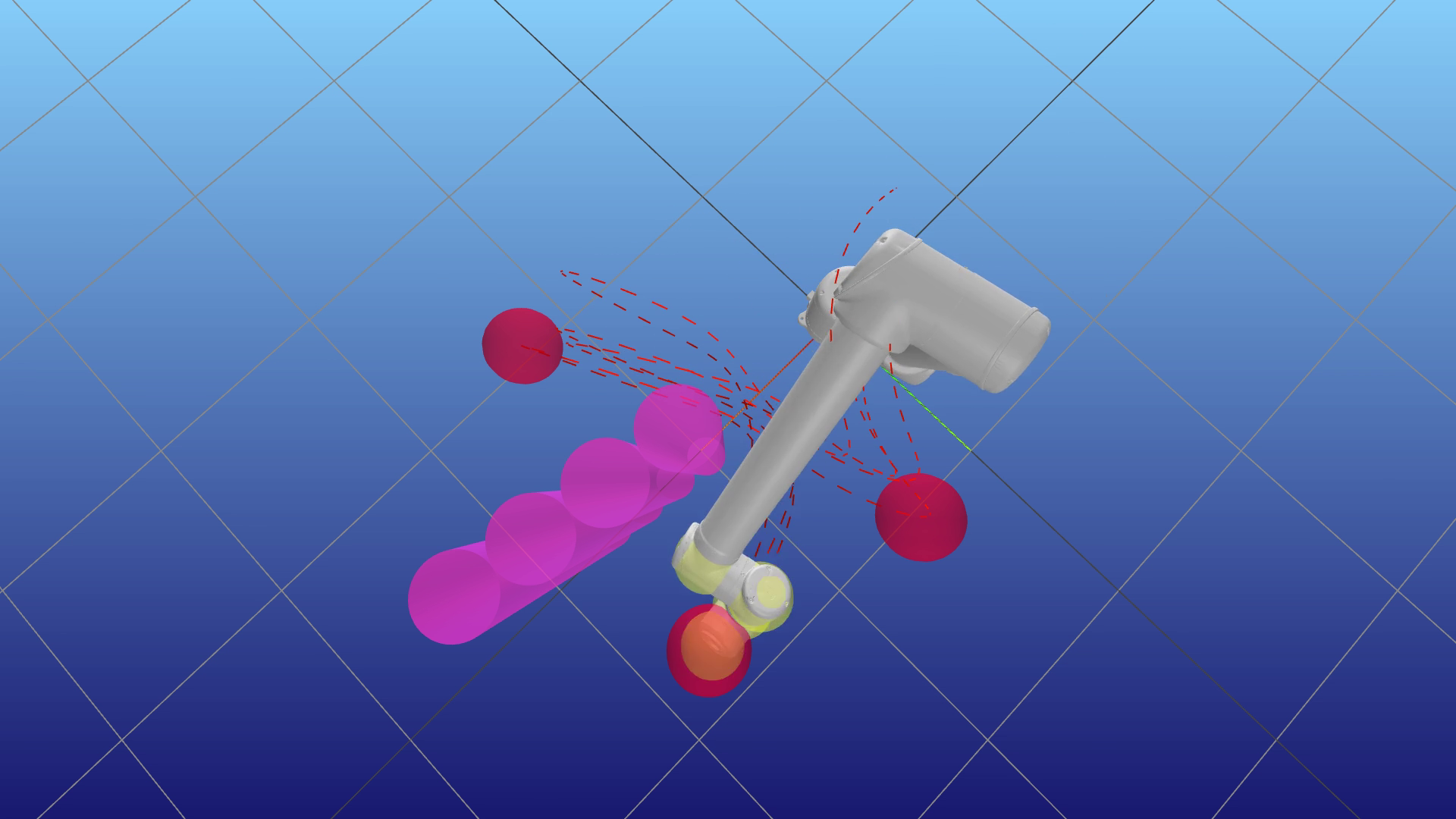}
    \end{tabular}
    \caption{\textbf{Left.} Throwing motion on UR10. The ball is the yellow, then green sphere. \textbf{Right.} UR10 reach task. The yellow spheres around the end-effector and wrist links do not collide with the purple cylinders, and the waypoints are reached at the specified times.}
    \label{fig:ur10_motions}
    \vspace*{-4mm}
\end{figure*}

\subsection{Backward and forward passes}

As outlined in the previous section, the semi-smooth (quasi-)Newton descent direction for $Q^l$ can be recovered from the system of equations:
\begin{equation}\label{eq:DDP_KKT}
	\calK_\mu
	\begin{bmatrix}
		\delta u \\ \delta x' \\ \delta\lambda \\
		[\delta\nu]_\calA
	\end{bmatrix}
	=-\begin{bmatrix}
		Q_u + Q_{ux}\delta x  \\
		Q_{x'} + Q_{x'x}\delta x \\
		f + f_x\delta x +  \mueq (\lambda_l - \lambda)  \\
		[h + h_x\delta x + \muin (\nu_l - \nu)]_\calA
	\end{bmatrix},
\end{equation}
where:
\begin{equation}\label{eq:DDP_regKKTmat}
	\calK_\mu \defeq
	\begin{bmatrix}
		Q_{uu}	& Q_{ux'}	& f_u^\top		& [h_u]_\calA^\top	\\
		Q_{x'u}	& Q_{x'x'}	& f_{x'}^\top	& [h_{x'}]_\calA^\top	\\
		f_u		& f_{x'}	& -\mueq I	& 			\\
		[h_u]_\calA			& [h_{x'}]_\calA	&	& -\muin I
	\end{bmatrix},
\end{equation}
is a regularized KKT matrix. It is similar to the matrix derived in \cite{jalletImplicitDifferentialDynamic2022}, with an additional block covering the active set of inequality constraints, denoted by $[\cdot]_\calA$.
Subscripted symbols (e.g. $f_x, h_u$\ldots) denote partial derivatives. 
We switch convention from \cite{jalletImplicitDifferentialDynamic2022}, where $\mu$ is the reciprocal of the parameters $(\mueq, \muin)$ we use here.

Since the previous state deviation $\delta x$ is unknown but the \textit{r.h.s.} of \eqref{eq:DDP_KKT} is linear in that parameter, we can recover the solution from the sensitivities, which satisfy:
\begin{equation}\label{eq:DDP_sensitivities}
    \calK_\mu
    \begin{bmatrix}
        k & K \\ a & A \\ \xi & \Xi \\
        \zeta_{\calA} & Z_{\calA}
    \end{bmatrix} =
    -\begin{bmatrix}
        Q_u     & Q_{ux}  \\
        Q_{x'}  & Q_{x'x} \\
        f + \mueq(\lambda_e-\lambda) & f_x \\
        [h + \muin(\nu_l-\nu)]_\calA & [h_x]_\calA
    \end{bmatrix}.
\end{equation}
The step is recovered as:
\begin{equation}\label{eq:pdDDP_step}
\begin{aligned}
    \delta u &= k + K\delta x, \quad \delta x' = a + A\delta x \\
    \delta\lambda &= \xi + \Xi\delta x, \quad
    \delta\nu = -{[\nu]}_{\calA^c} + \zeta_\calA + Z_\calA\delta x.
\end{aligned}
\end{equation}
In practice, the system \eqref{eq:DDP_sensitivities} is solved by an $LDL^\top$ Cholesky factorization of the KKT matrix $\calK_\mu$.

\noindent
\textbf{Forward pass and linear rollout.} Similarly to \cite{jalletImplicitDifferentialDynamic2022}, the primal-dual step is recovered by a linear rollout over \eqref{eq:pdDDP_step}:
\begin{equation}
\begin{aligned}
    \delta u_t &= k_t + K_t\delta x_t,\
    \delta x_{t+1} = a_t + A_t\delta x_t \\
    \delta\lambda_{t+1} &= \xi_t + \Xi_t\delta x_t,
    \ 
    \delta \nu_{t+1} = -[\nu_{t+1}]_{\calA^c} + \zeta_{\calA,t} + Z_{\calA,t}\delta x_t.
\end{aligned}
\end{equation}
The initial step over $(\delta x_0, \delta\lambda_0, \delta\nu_0)$ is associated with the value function, equality and inequality constraints at $k=0$.

\subsection{Convergence and globalization strategy}

As discussed in Sec.~\ref{sec:augmented_lagrangian} and following the approach proposed in~\cite{kazdadiEqualityConstrainedDifferential2021,jalletImplicitDifferentialDynamic2022}, we use a BCL strategy as an outer loop to automatically update the parameters $\mu$ and $\rho$ and the multipliers estimates $(\lambda_l, \nu_l)$ according to the progress made on the primal and dual feasibility. We refer to~\cite{jalletImplicitDifferentialDynamic2022} to see how BCL is used within the constrained DDP.
We also use a backtracking line-search procedure to compute a step length $\alpha > 0$ at each iteration of the algorithm after the linear rollout.
This line-search relies on the assumption that the direction $\delta\bfm{w} = (\delta\bfx, \delta\bfu,\delta\bmlam, \delta\bmnu)$ is a descent direction satisfying $\nabla\calM_{\mu,\rho}^\top \delta\bfm{w} < 0$. Denoting \mbox{$\phi(\alpha) = \calM_{\mu,\rho}(\bfm{w}+\alpha \delta\bfm{w};\bfm{w}_l)$}, our Armijo backtracking procedure looks for the first $k \in \NN$ such that $\phi(t^k) \leq \phi(0) +c_1 t^k\phi'(0)$ and sets $\alpha = t^k$.
To ensure the descent condition, we play on the proximal parameters $\rho_l > 0$ in the outer BCL loop, and have a heuristic similar to \cite{wachterImplementationInteriorpointFilter2006} and \cite{tassaSynthesisStabilizationComplex2012} to control the inertias of the regularized KKT matrices \eqref{eq:DDP_regKKTmat}, which is central to obtain good convergence behavior.
Our stopping criteria is the same as in the constrained optimization framework outlined in Alg.~\ref{al:PDAL_BCL}.

%% file: parts/experiments.tex
\sisetup{exponent-product=.}

\section{Experiments}
\label{sec:experiments}

For experimental validation our approach, we extend the numerical optimal control framework of~\cite{jalletImplicitDifferentialDynamic2022}, written in Python, which relies on the Pinocchio rigid-body dynamics library \cite{carpentierPinocchioLibraryFast2019} for providing analytical derivatives~\cite{carpentierAnalyticalDerivativesRigid2018} and NumPy \cite{harrisArrayProgrammingNumPy2020} for linear algebra.
Because it is in Python, we do not provide CPU timings against existing implementations.

\subsection{Bound-constrained problems}

\noindent
\textbf{Bound-constrained LQR.}
The first system we test is the simple linear-quadratic regulator (LQR) with bound constraints, of the form:
\begin{equation}\label{eq:boundLQR}
\begin{alignedat}{3}
	\min_{\bfx, \bfu} &\quad  \sum_{k=0}^{N-1} \frac{1}{2}x_k^\top Qx_k + \frac{1}{2}u_k^\top R u_k
	+ \frac{1}{2} x_N^\top Q_N x_N  \\
	\suchthat	& x_{k+1} = Ax_k + Bu_k + c, \ k \leq N-1 \\
				& x_0 = \bar{x}_0,\ {-\bar{u} \leq u_k \leq \bar{u} }
\end{alignedat}
\end{equation}
where $Q, Q_N \text{ and } R$ are positive semi-definite matrices, \mbox{$\bar{u} \in \overline{\RR}_+^{n_u}$} are the control bounds. 
This problem is a convex quadratic program (QP), which can be solved with classical QP solvers.
We test a few configurations for the problem parameters $(A, c, \bar{u})$, leading to the results in Fig.\,\ref{fig:boundLQR_1} and \ref{fig:boundLQR_2}.
For bound-constrained LQR, the proposed method takes about ten iterations to converge to an optimal solution with precision $\epsilon = 10^{-8}$.

\def\figwidth{0.8\linewidth}

\begin{figure}[ht!]
    \vspace{2mm}
	\centering
	\includegraphics[trim=0 10pt 0 40pt,clip=true,width=\figwidth]{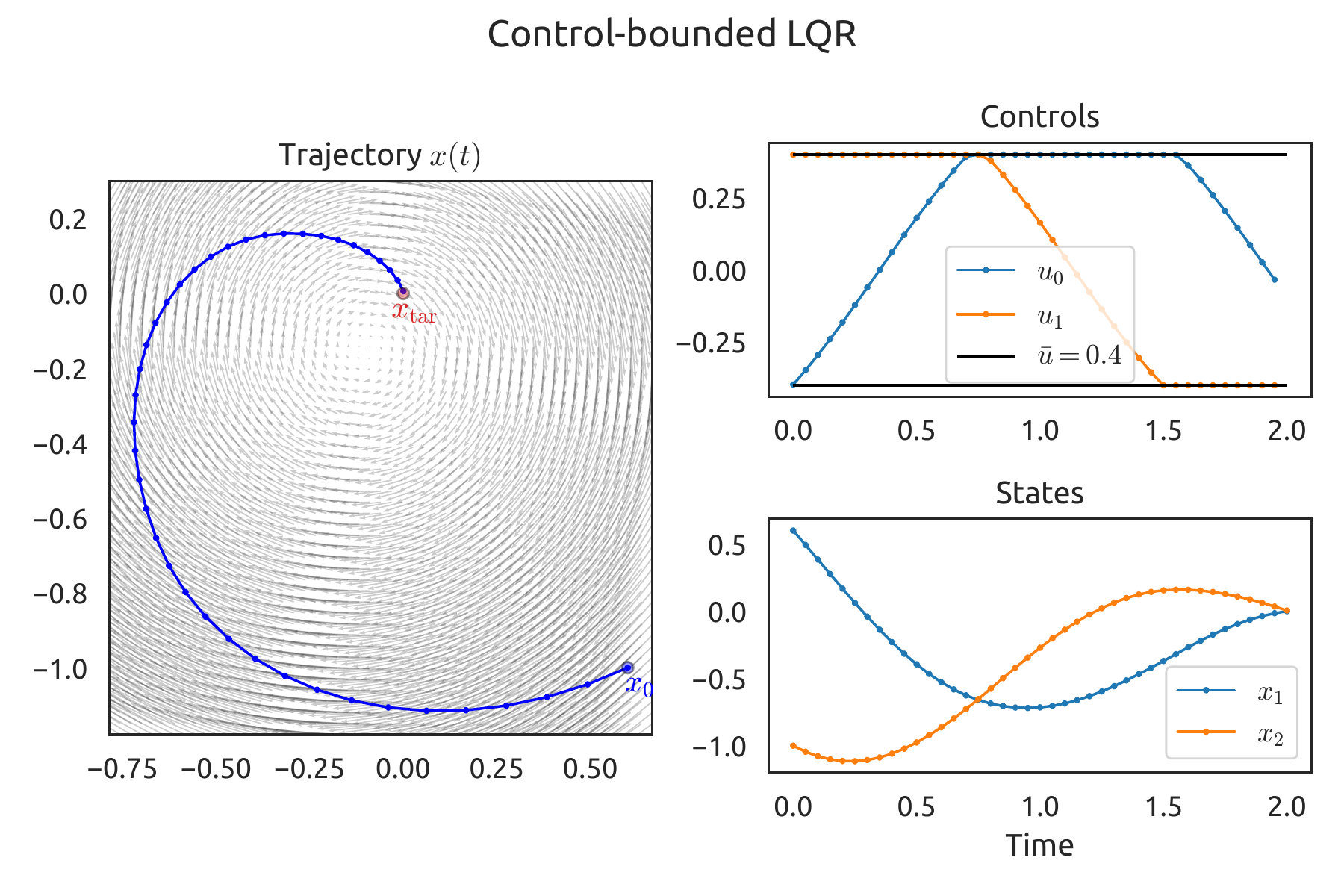}
	\caption{Rotational system: $A$ approximates a continuous-time system with matrix $A_c = \begin{bsmallmatrix}0 & 2 \\ -2 & 0
		\end{bsmallmatrix}$, $c=(0.3, -0.2)$, control bound $\bar{u}=0.4$. In spite of the control bounds, which saturate, the target is reached.}
	\label{fig:boundLQR_1}
	\vspace*{-4mm}
\end{figure}

\begin{figure}[ht!]
	\centering
	\includegraphics[trim=0 10pt 0 40pt,clip=true,width=\figwidth]{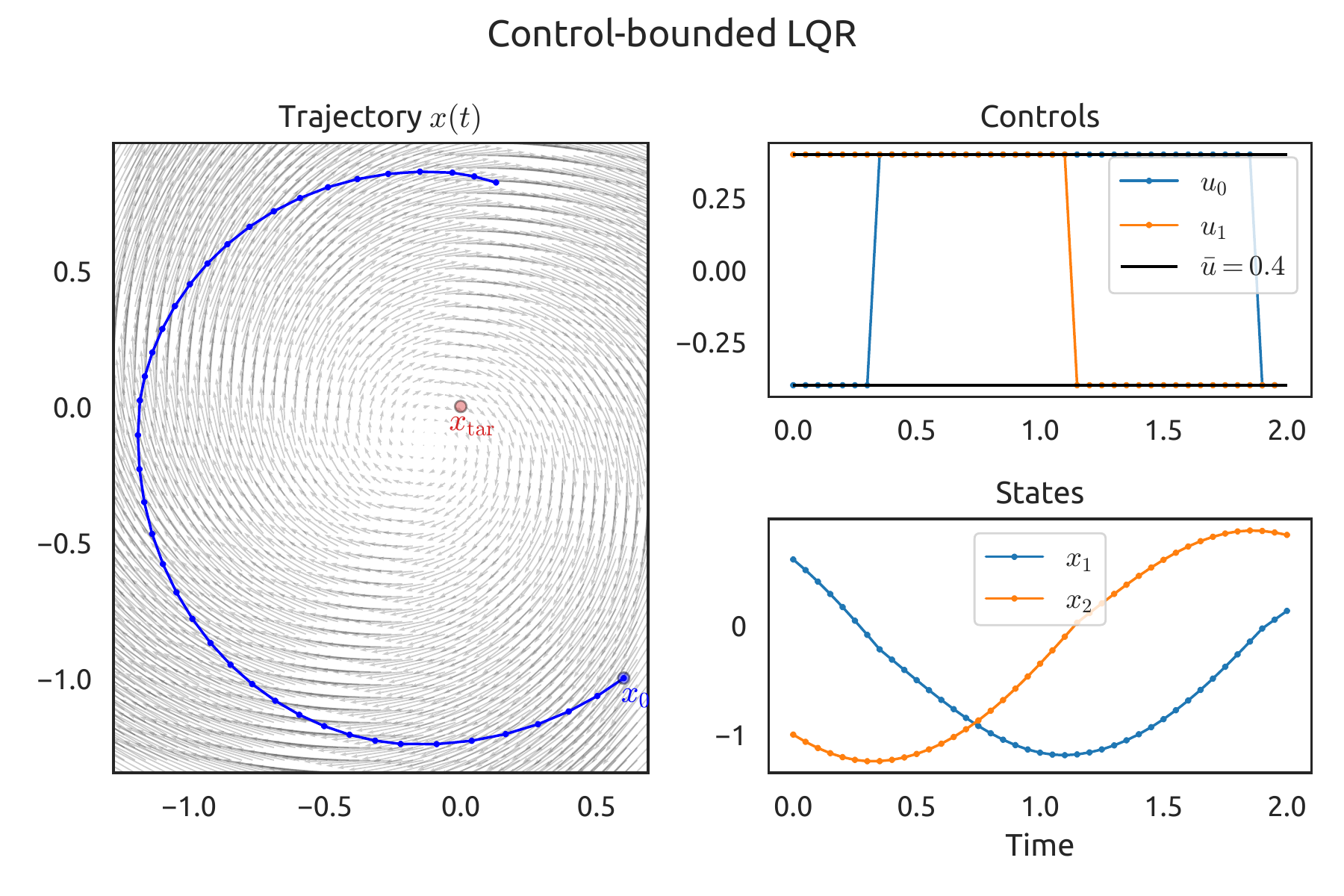}
	\caption{The continuous time system is $A_c = \begin{bsmallmatrix}0.4 & 2 \\ -2 & 0.4\end{bsmallmatrix}$, $c$ is the same, the target is located near a repulsive equilibrium. Even with the same bound $\bar{u}=0.4$, the target is not reached. The obtained control is a bang-bang control.}\label{fig:boundLQR_2}
	\vspace*{-4mm}
\end{figure}

\begin{figure}[ht!]
    \centering
    \includegraphics[trim=10pt 10pt 5pt 10pt,clip=true,width=.9\linewidth]{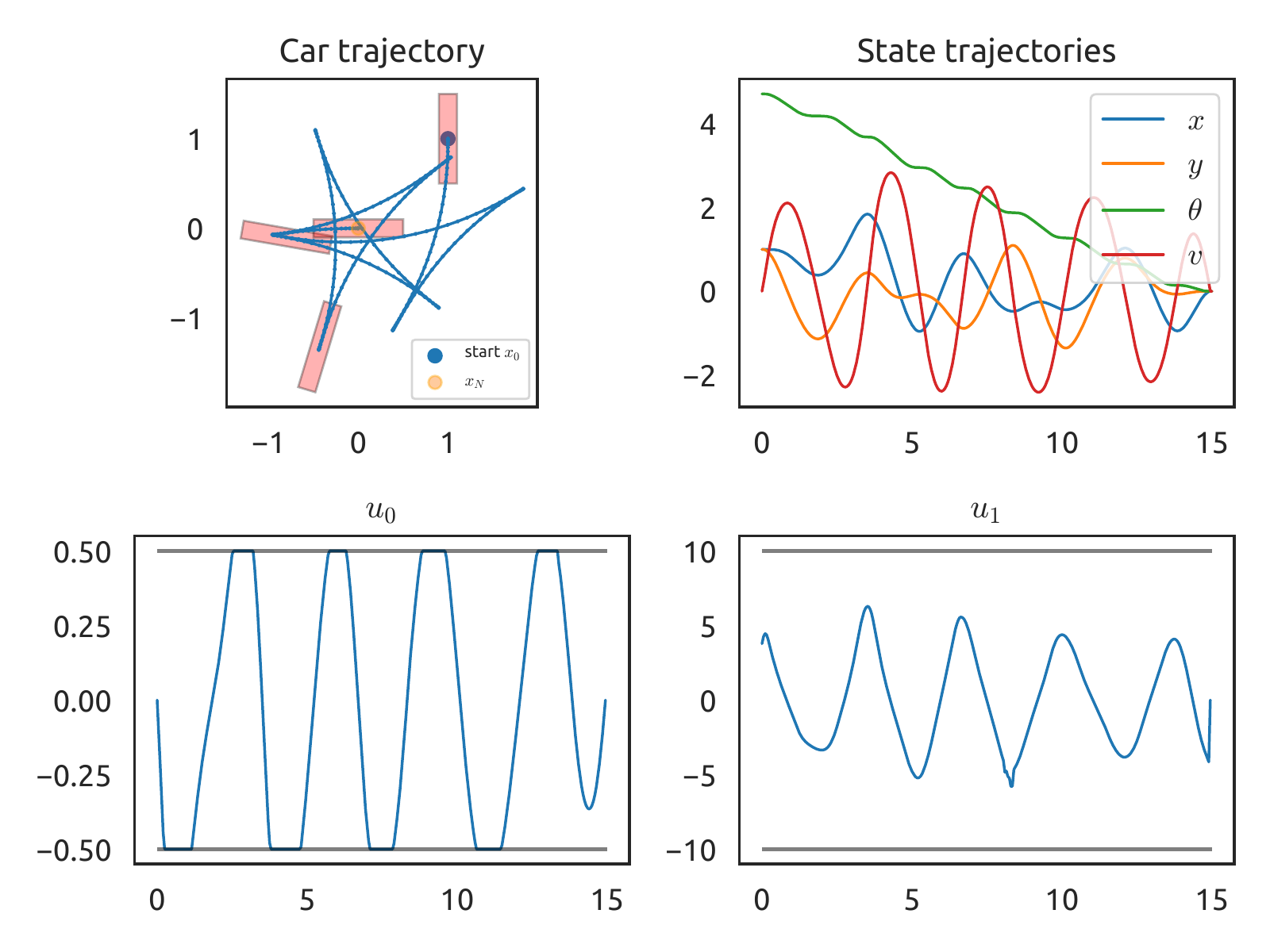}
    \caption{Solution of the car parking task. The starting state is $(1, 1, \frac{3\pi}{2}, 0)$. The turn control $u_0=\omega$ often saturates, causing the policy to go backwards to turn further; due to the parametrization with angle $\theta$, a lot more turning is required.}
    \label{fig:carparking:traj}
\end{figure}

\noindent
\textbf{Car parking}, as proposed in~\cite[IV.B.]{tassaControllimitedDifferentialDynamic2014}.
The car dynamics is defined by its state variables $(x,y,\theta,v)$, the goal is to steer the car to the state $(0,0,0,0)$. 
The control inputs are the front wheel acceleration $a\in\RR$ and angle $\omega$, with bounds $|a| \leq \qty{10}{\meter\per\second^2},|\omega| \leq \qty{0.5}{\per\second}$.
Figure\,\ref{fig:carparking:traj} illustrates the resulting trajectory with comments.
Following~\cite{tassaControllimitedDifferentialDynamic2014}, the system makes the distinction between the initial angles $\frac{3\pi}{2}$ and $-\frac{\pi}{2}$, which (interestingly for the interest of the benchmark) forces the solver to find more commutations. 
We used initial penalty parameters $\mu_0 = 100,\rho_0=10^{-5}$, and convergence threshold $\epsilon = \num{2e-4}$ (no convergence threshold was given in \cite{tassaControllimitedDifferentialDynamic2014}). 
The timestep is $dt=\qty{0.03}{\second}$ and horizon $T=\qty{15}{\second}$.
The problem converges to an optimal solution in 62 iterations.

\begin{figure}[ht!]
    \vspace*{-2mm}
    \centering
    \includegraphics[trim=0 0 0 0,clip=true,width=1\linewidth]{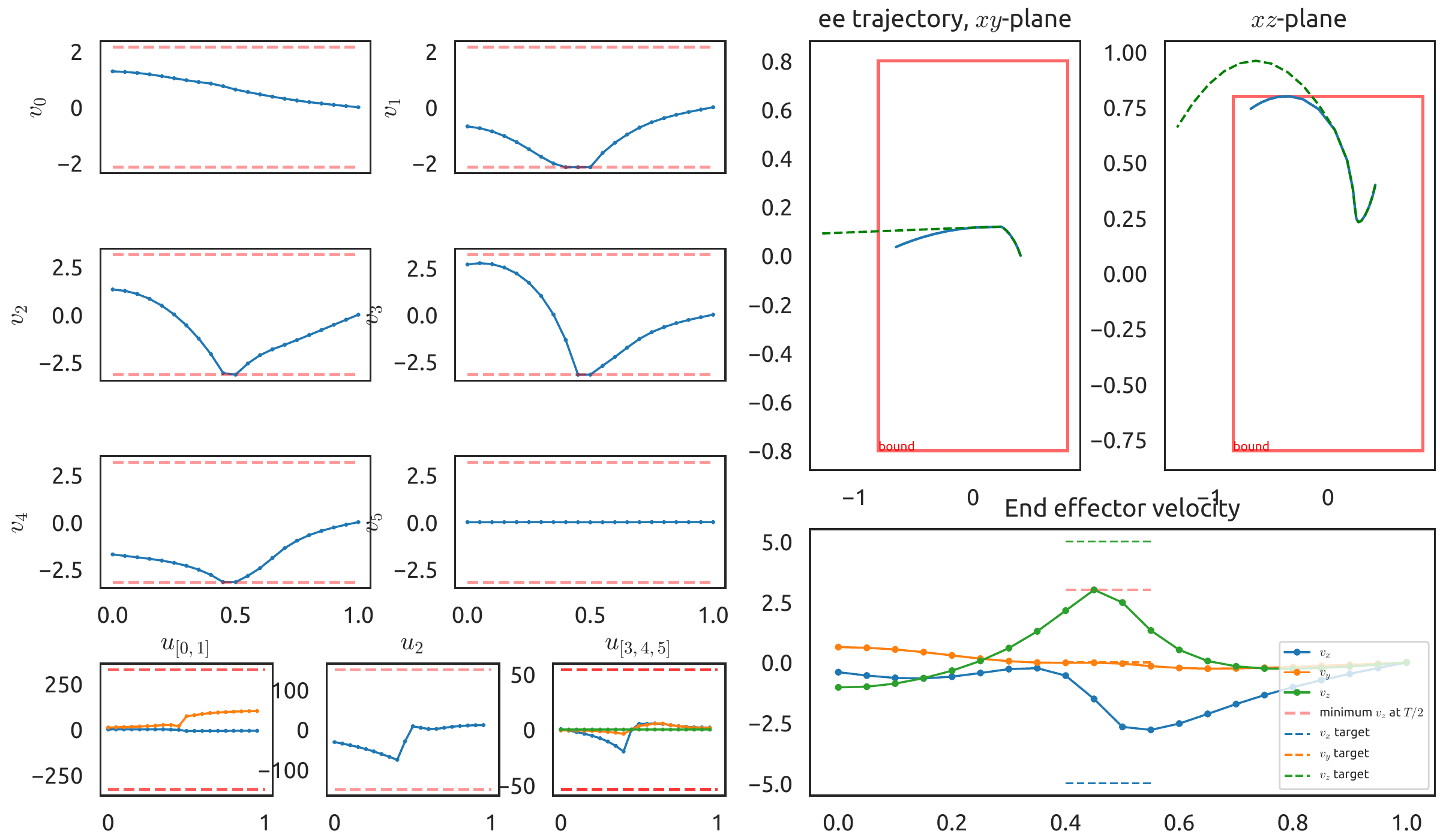}
    \caption{Joint angles (upper left), joint velocities (middle left), controls (lower left), end-effector trajectory (upper right) and velocity (lower right) for the throwing motion on UR10. The ball trajectory is displayed with dashed green lines on the upper-right plots. Both the velocities, controls and end-effector position satisfy their respective bounds, displayed in \textcolor{red!60!Black}{red}. The minimum target end-effector velocity is also satisfied.}
    \label{fig:ur10_throw_plots}
    \vspace*{-6mm}
\end{figure}

\noindent
\textbf{UR5 -- throwing task.} The goal of this task is to throw a ball at a target velocity $\bar{v}$ at a time half-way through the horizon $T = \qty{1}{\second}$ (with a minimum velocity in the $z$ direction). We constrain the elbow frame to be above ground, the end-effector stay within a box, along with joint velocity and torque limits.
The dynamics are integrated using a second order Runge-Kutta scheme with timestep $dt=\qty{0.05}{\second}$.
The state and control trajectories satisfy the bound constraints as depicted in  in Fig.~\ref{fig:ur10_throw_plots}. 
The robot motion is illustrated in Fig.~\ref{fig:ur10_motions}.

\subsection{Obstacle-avoidance}

\begin{figure}[ht!]
	\centering
	\includegraphics[trim=15pt 10pt 0 0,clip=true,width=.75\linewidth]{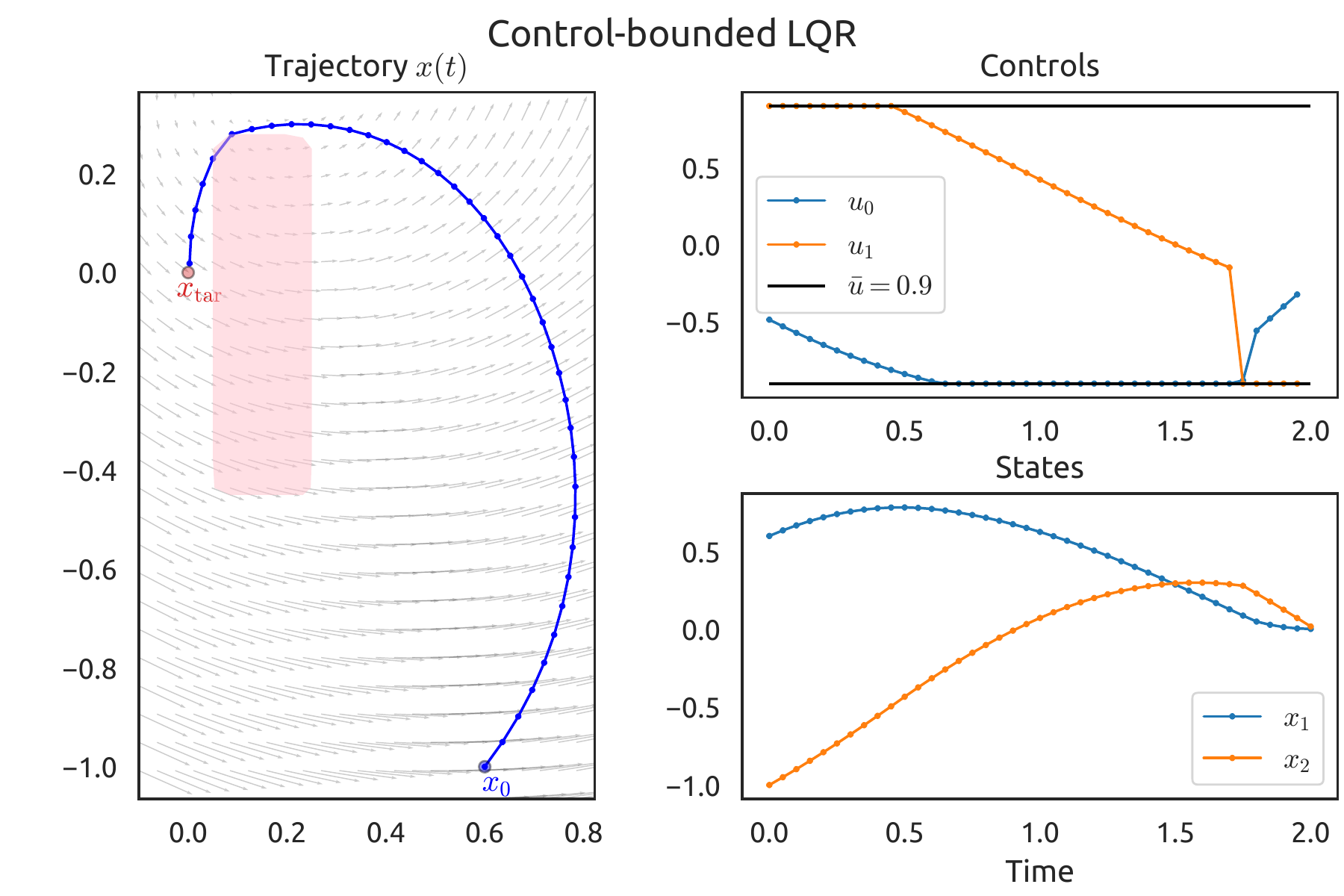}\\
	\includegraphics[trim=10pt 10pt 10pt 0,clip=true,width=.75\linewidth]{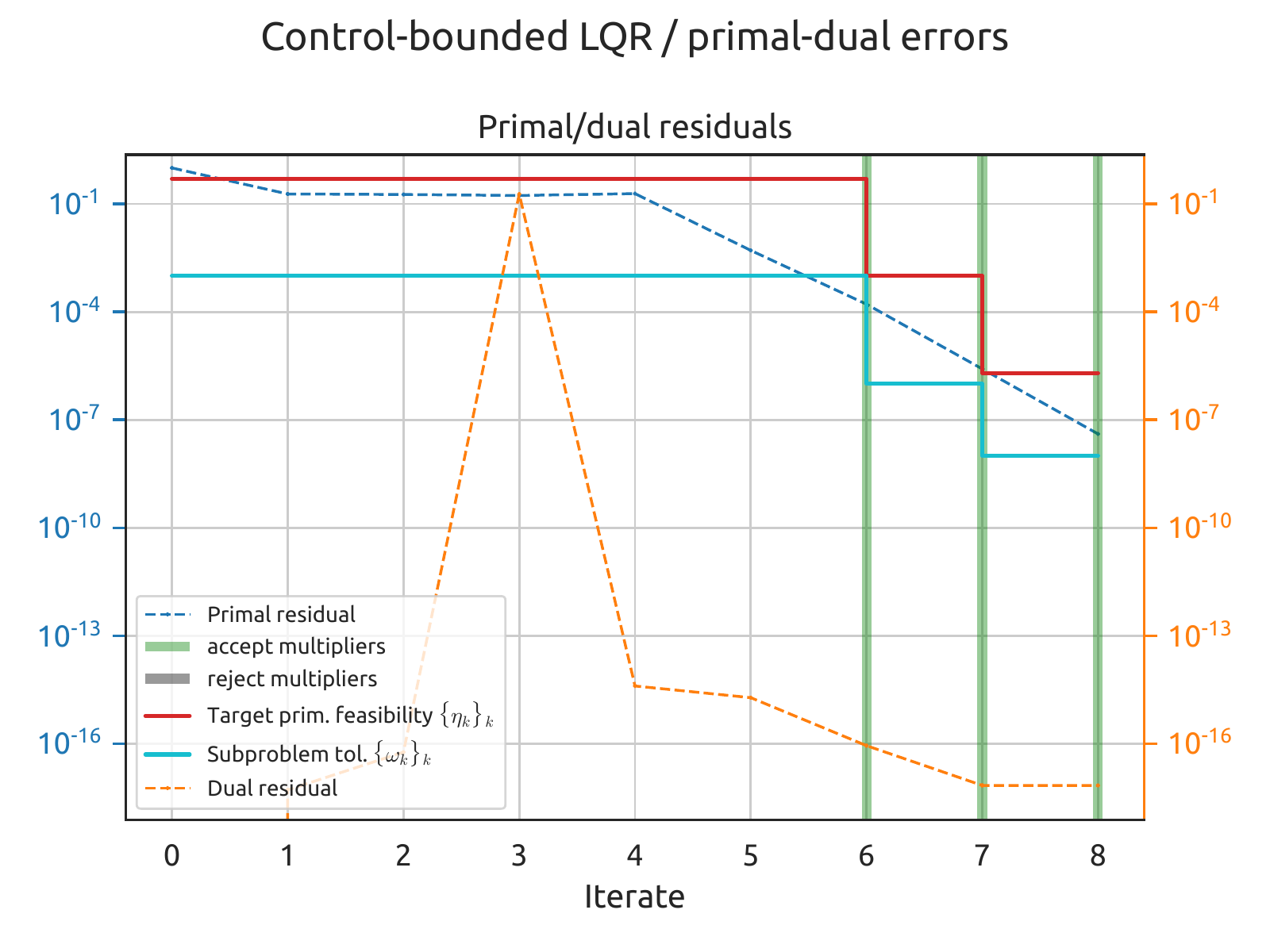}
	\vspace*{-2mm}
	\caption{\textbf{Top:} The pink area is a rectangular obstacle defined by \eqref{eq:polyhedralAvoid}. 
	The trajectory avoids the area (at the discretization nodes) and the controls saturate: both constraints are satisfied. 
	\textbf{Bottom:} Evolution of the primal-dual residuals after each step (backward and forward pass). We obtain very fast convergence in a handful of steps.}
	\label{fig:boundLQR_obstacle1}
	\vspace*{-2mm}
\end{figure}

\noindent
\textbf{LQR with obstacles.}
We extend the exemple of the bound-constrained LQR \eqref{eq:boundLQR} with path constraints that consist in avoiding obstacles. We consider avoiding the interiors of polyhedral sets of the form $P^{(j)} = \{ x \mid C^{(j)} x \leq d^{(j)} \}$ which is the piecewise linear constraint
\begin{equation}\label{eq:polyhedralAvoid}
	\max_i{} (C^{(j)} x - d^{(j)})_i \geq 0.
\end{equation}
These constraints make the problem nonconvex and thus cannot be handled by standard convex solvers. Fig.\,\ref{fig:boundLQR_obstacle1} shows an example with both obstacles and control which saturate.

\newcommand{\dist}{\mathrm{dist}}

\begin{figure}[ht!]
    \vspace{2mm}
    \centering
    \includegraphics[trim=10pt 10pt 0 0,clip=true,width=.96\linewidth]{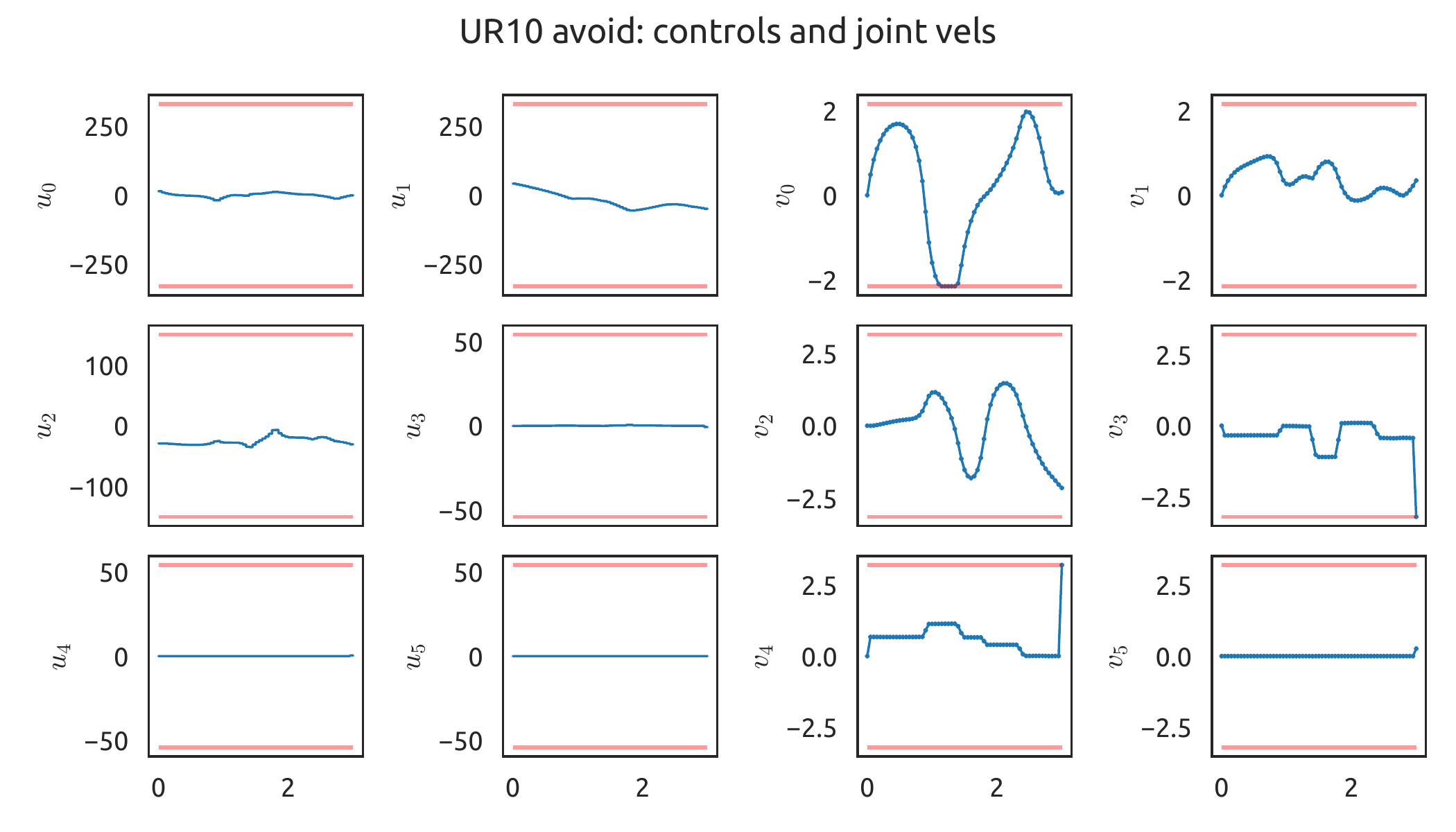}%
    \caption{Controls and velocities for the UR10 reach task. Translucent red lines indicate control and velocity bounds. The velocities saturate only for a few axes in the middle and end of the trajectory (the arm bows down to avoid the obstacles, and lurches forward to reach the final waypoint).}\label{fig:ur10_reach_controls}
    \vspace*{-6mm}
\end{figure}

\noindent
\textbf{UR10 -- reach task with obstacles.} 
The goal is for the end-effector $p_e(q)$ to reach a target $\bar{p} \in \RR^3$, expressed as a terminal cost $\ell_\rmf(x) = \frac{1}{2}\| p_e(q) - \bar{p} \|_{W_\mathrm{ee}}^2$.
We also impose waypoint constraints at $t_0, t_1 \in (0,T)$, with time horizon $T=\qty{3}{\second}$.
As obstacles, we choose simple vertical cylinders of radius $r_C$ and impose that they should not collide with spheres of radius $r_S$ centered around given frames $p_j$ (the end-effector and wrist links of the UR10). This condition is expressed using the distance from the sphere center $p_j$ to the cylinder axis: $\|p_j - \proj_{\text{cyl. axis}}(p_j)\| \geq r_S + r_C$.
Figure\,\ref{fig:ur10_motions} illustrates the motion on the UR10 robot, and Fig.\,\ref{fig:ur10_reach_controls} controls and velocities.

%% file: parts/conclusion.tex
\section{Conclusion}
\label{sec:conclusion}

In this work, we have introduced a new approach for solving generic NLPs with equality and inequality constraints.
We propose combining the BCL globalization strategy~\cite{connGloballyConvergentAugmented1991} with the minimization of a relaxed semi-smooth primal-dual Augmented Lagrangian function inspired by~\cite{gillPrimaldualAugmentedLagrangian2012}. 
We then apply this approach to extend the framework of equality-constrained \cite{kazdadiEqualityConstrainedDifferential2021} and dynamics-implicit \cite{jalletImplicitDifferentialDynamic2022} DDP to the case of inequality constraints.
It results in an overall second-order quasi-Newton-like algorithm for solving constrained DDP problems. 
We finally highlight the numerical efficiency of our method on various sets of standard case-studies from the robotic literature.
These contributions pave the way towards more advanced numerical methods for dealing with complex optimization problems in robotics, with the ambition of significantly reducing the computational burden, increase the numerical robustness of the trajectory optimization methods while also lowering the need of manually tuning underlying hyper-parameters. %
As future work, we plan to implement our contributions in C++ within the Crocoddyl library~\cite{mastalliCrocoddylEfficientVersatile2020}, to properly account for equality and inequality constraints in trajectory optimization.